\documentclass[11pt]{article}

\usepackage[preprint]{acl}

\usepackage{times}
\usepackage{latexsym}
\usepackage[utf8]{inputenc}
\usepackage{csquotes}
\usepackage{graphicx}
\usepackage{multirow}
\usepackage{booktabs}
\usepackage{makecell}
\usepackage{amsmath} 
\usepackage{amssymb} 

\usepackage[T1]{fontenc}

\usepackage[utf8]{inputenc}

\usepackage{microtype}

\usepackage{inconsolata}

\usepackage{graphicx}
\usepackage[most]{tcolorbox}

\usepackage{subcaption}

\usepackage{hyperref}

\title{Making Implicit Preservation Intent Explicit in \\ Conversational Image Editing}

\author{
 \textbf{Soomin Han\textsuperscript{1,$\dagger$}},
 \textbf{Jihyung Ahn\textsuperscript{2,$\dagger$}},
 \textbf{Bumsoo Kim\textsuperscript{3,*}},
 \textbf{Buru Chang\textsuperscript{2,*}}
\\
 \textsuperscript{1}Sogang University,
 \textsuperscript{2}Korea University,
 \textsuperscript{3}Chung-Ang University
\\
 \texttt{soominsion@sogang.ac.kr},
 \texttt{gina0520@korea.ac.kr},
 \texttt{bumsoo@cau.ac.kr},
 \texttt{buru\_chang@korea.ac.kr}
\\
 \textsuperscript{$\dagger$}Equal contribution.
 \textsuperscript{*}Corresponding authors.
}

\begin{document}
\maketitle

\begin{abstract}
Conversational image editing requires preserving not only visible content, but also content that temporarily disappears across turns.
When newly added or modified content occludes a previously visible region, that region should reappear if it was never semantically changed.
However, existing systems often fail to recover such occluded-but-unchanged content, producing inconsistent or hallucinated results.
We introduce \textbf{\textit{OCCUR-Bench}}, a diagnostic benchmark for temporal preservation in conversational image editing.
OCCUR-Bench provides diverse occlusion-and-revelation scenarios with historical restoration references, enabling evaluation of faithful restoration rather than plausible regeneration.
We also propose \textbf{\textit{ReSpec}}, a training-free framework that makes implicit preservation explicit by pairing restoration-aware instructions with historical visual references.
Given an editing history, ReSpec identifies what should persist, selects the historical image state that provides missing visual evidence, and conditions an in-context editor on the resulting instruction and reference image.
Experiments show that ReSpec improves restoration fidelity and temporal consistency on OCCUR-Bench, highlighting the need to ground preservation in editing history rather than only the current image. The dataset and code are available at \url{https://github.com/anonymous745961852-cloud/implicit-preservation-editing}.
\end{abstract}
\section{Introduction}
\label{sec:introduction}

\textit{Conversational image editing} enables users to refine visual content through iterative dialogue.
At each turn, the user requests an edit, observes the updated image, and provides a follow-up instruction.
Unlike single-turn editing, each instruction must therefore be interpreted in the context of prior edits, intermediate results, and user expectations accumulated across turns.
A basic expectation is \textit{preservation}: content that the user has not asked to change should remain consistent.

\begin{figure*}[t]
    \centering
   \includegraphics[width=\textwidth]{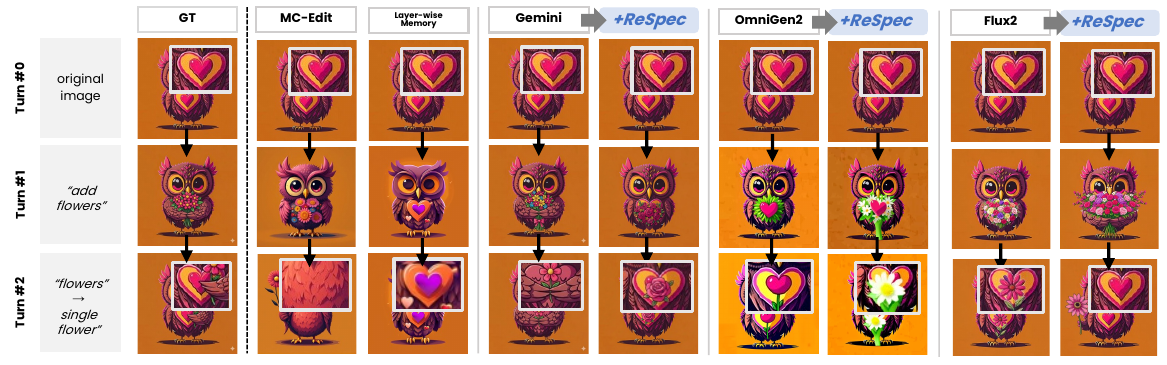} 
\caption{Motivating example of temporal preservation in conversational image editing.
The owl's chest pattern is occluded by added flowers but never semantically modified, so it should be restored when revealed.
Existing systems often regenerate the revealed region inconsistently, whereas our proposed ReSpec grounds preservation in the editing history and better recovers the original pattern.}  \label{fig:1_motivating_results}
    \vspace*{-0.5em}
\end{figure*}

Existing image editing methods are primarily designed to preserve what remains visible in the current image.
However, conversational editing also requires preserving what temporarily disappears.
Such cases naturally arise in iterative editing: common operations such as adding, moving, resizing, replacing, or stylizing objects can temporarily hide previously visible content and reveal it again in later turns.
As illustrated in Figure~\ref{fig:1_motivating_results}, flowers added over an owl's chest can hide the original feather texture.
When the flowers are later removed or resized, the hidden texture should reappear because it was never semantically modified.
Current conversational editing systems can fail in this setting, producing inconsistent or hallucinated content instead.
Once the texture is occluded, the current image no longer provides visual evidence of its original appearance.
This reveals a simple but important limitation: visual absence does not imply semantic change.

This failure mode is not well captured by existing conversational image editing benchmarks.
Most evaluations focus on instruction following, visible-region preservation, or final-output quality.
They therefore do not directly test whether a model can recover content that was visible in an earlier turn, temporarily occluded, and later revealed.
As a result, a model may appear successful under standard evaluation while still failing to preserve content that should persist across the editing history.

To evaluate this problem, we introduce \textbf{\textit{OCCUR-Bench}} (\underline{O}cclusion \underline{C}onsistency and \underline{C}ontent \underline{U}nveiling for \underline{R}estoration), a diagnostic benchmark for temporal preservation in conversational image editing.
OCCUR-Bench simulates this recurring pattern through diverse occlusion-and-revelation scenarios built from common multi-turn editing operations.
Each scenario first occludes existing content without semantically modifying it, may apply intermediate edits, and later reveals the occluded region through operations such as removal, movement, resizing, or replacement.
Each scenario includes a historical reference state that contains the correct appearance of the revealed content, allowing us to distinguish faithful restoration from plausible hallucination.

Using OCCUR-Bench, we find that existing conversational editing systems struggle to recover occluded-but-unchanged content.
These systems primarily respond to the explicit instruction at each turn; preservation is left to the visual evidence available in the current image.
When unchanged content becomes occluded, it is neither specified by the instruction nor available in the current image.
We therefore propose ReSpec, a training-free framework that makes implicit preservation explicit.
As shown in Figure~\ref{fig:1_motivating_results}, applying ReSpec helps the editor recover the owl's original chest pattern rather than regenerating the revealed region from the current image alone.
Given an editing history, ReSpec identifies the content that should persist, expresses this preservation target in a restoration-aware instruction, and supplies the missing visual evidence by selecting a historical reference image where the content is visible and valid.

Experiments on OCCUR-Bench show that ReSpec improves restoration fidelity and temporal consistency when paired with in-context image editing models that accept reference images.
Trajectory-length analysis further shows the importance of historical reference grounding when hidden content must be restored across multiple turns.
These results suggest that robust conversational image editing requires moving beyond preserving only what is currently visible, toward explicitly grounding preservation in the editing history.

Our contributions are summarized as follows:
\begin{enumerate}
    \item We identify \textit{temporal preservation} as a key challenge in conversational image editing, where unchanged content may be temporarily occluded but still expected to persist.

    \item We introduce \textbf{\textit{OCCUR-Bench}}, a diagnostic benchmark for occlusion-and-revelation scenarios, and show that existing conversational editing systems struggle to restore occluded-but-unchanged content.

    \item We propose \textbf{\textit{ReSpec}}, a training-free framework that makes implicit preservation explicit through restoration-aware instructions and historical visual references.
\end{enumerate}
\section{Preliminaries}
\label{sec:preliminary}

\subsection{Conversational Image Editing}
\label{subsec:prelim_cie}

We consider a conversational image editing setting in which an image is iteratively modified through a sequence of natural language instructions.
Let \(I_0\) denote the base image and \(P_t\) denote the editing instruction at turn \(t\).
Given the editing history
\begin{equation}
\mathcal{H}_{t-1}
=
\{I_0,P_1,I_1,P_2,\dots,I_{t-2},P_{t-1}, I_{t-1}\},
\label{eq:history}
\end{equation}
a conversational editing model generates the updated image:
\begin{equation}
I_t
=
f_{\mathrm{edit}}(\mathcal{H}_{t-1},P_t),
\label{eq:editing}
\end{equation}
where \(f_{\mathrm{edit}}(\cdot)\) denotes the editing model.
This formulation captures the sequential nature of conversational editing: the current output may depend not only on the current instruction, but also on previous instructions and intermediate image states.

\subsection{Temporal Occlusion}
\label{subsec:temporal_occlusion}

Existing conversational editing methods often assume that preservation can be determined from the currently visible image.
This assumption becomes insufficient under temporary occlusion, where newly generated or transformed content visually covers existing content.

To reason about occlusion, let \(\mathcal{O}=\{o_1,\dots,o_N\}\) denote the set of semantic scene entities.
We distinguish between the observed image and a conceptual latent scene state that represents the semantic content of the scene.
Let
\begin{equation}
S_t
=
\{o_i^{(t)}\}_{i=1}^{N}
\label{eq:scene_state}
\end{equation}
denote the latent scene state at turn \(t\), where \(o_i^{(t)}\) denotes the semantic state of entity \(o_i\).
Here, \(S_t\) is a conceptual abstraction rather than an explicitly estimated representation.
The observed image is viewed as a rendering of this latent state:
\begin{equation}
I_t
=
\mathcal{R}(S_t),
\label{eq:rendering}
\end{equation}
where \(\mathcal{R}(\cdot)\) denotes the image formation operator.

Temporal occlusion occurs when an entity remains part of the latent scene state but is not visible in the observed image:
\begin{equation}
o_i^{(t)} \in S_t,
\quad
\neg \mathrm{visible}(o_i^{(t)}, I_t).
\label{eq:latent_occlusion}
\end{equation}
In this case, the current image no longer provides direct visual evidence of the occluded entity.
At editing turn \(t\), the desired scene state after editing may therefore not be recoverable from the current input image and instruction alone:
\begin{equation}
p(S_t \mid I_{t-1},P_t)
\neq
p(S_t \mid \mathcal{H}_{t-1},P_t).
\label{eq:history_dependency}
\end{equation}
Temporal occlusion becomes a preservation problem when the occluded entity has not been semantically modified by the editing instructions.

\subsection{Temporal Preservation}
\label{subsec:temporal_preservation}

Temporal preservation concerns entities that remain semantically unchanged across the editing trajectory, regardless of whether they are currently visible.
To define this, let \(\mathcal{M}_t\subseteq\mathcal{O}\) denote the entities whose semantic state is explicitly modified by instruction \(P_t\) at turn \(t\).
We define the persistent entity set at turn \(t\) as
\begin{equation}
\mathcal{P}_t
=
\{o_i \in \mathcal{O}
\mid
o_i \notin \mathcal{M}_k,\ \forall k \le t
\}.
\label{eq:persistent_entities}
\end{equation}
%

Entities in \(\mathcal{P}_t\) should remain recoverable after editing, including cases where they become temporarily invisible due to occlusion.
Accordingly, a conversational editor should not only execute the current instruction, but also preserve unchanged visible content and restore persistent entities when they reappear after occlusion.
This intuition motivates OCCUR-Bench, which diagnoses failures of temporal restoration, and ReSpec, which grounds restoration in historical visual evidence.

\section{OCCUR-Bench}
\label{sec:occur_bench}

Existing conversational image editing benchmarks focus on instruction following and visible-region consistency, but do not assess whether temporally occluded content is faithfully restored after revelation.
To address this gap, we introduce \textbf{\textit{OCCUR-Bench}} (\underline{O}cclusion \underline{C}onsistency and \underline{C}ontent \underline{U}nveiling for \underline{R}estoration), a diagnostic benchmark for evaluating temporal preservation in conversational image editing.
OCCUR-Bench targets occlusion-and-revelation trajectories in which content becomes temporarily hidden, remains semantically unchanged, and must be restored when it becomes visible again.


\subsection{Scenario Design}
\label{subsec:scenario_design}

\begin{figure}[t]
    \centering
    \includegraphics[width=\columnwidth]{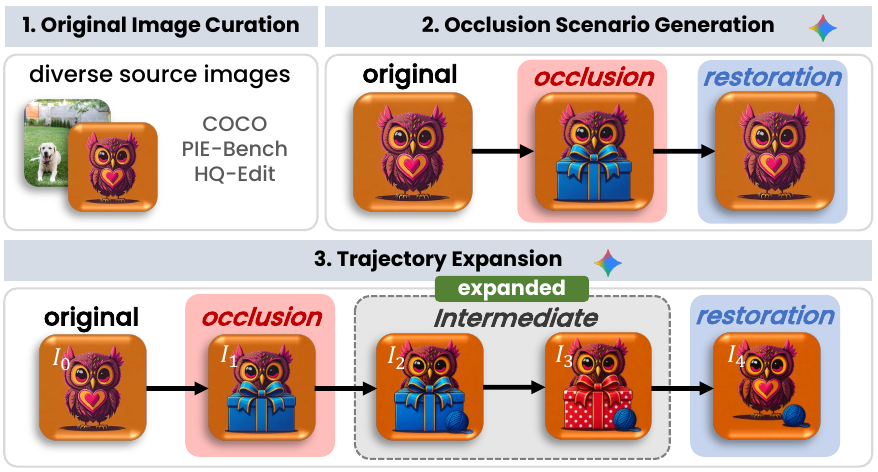}
\caption{
OCCUR-Bench scenario design.
Each trajectory temporarily occludes a semantically unchanged occludee, applies optional intermediate edits, and later reveals it by modifying the occluder.
Evaluation compares the revealed region with \(I_{\mathrm{ref}}\) to measure faithful restoration rather than plausible regeneration.
}   \label{fig:6_dataset_construction}
\vspace{-0.5em}
\end{figure}

Each OCCUR-Bench scenario is designed to isolate temporal preservation under occlusion-and-revelation, as illustrated in Figure~\ref{fig:6_dataset_construction}.
We refer to the newly introduced or transformed entity that hides existing content as the \emph{occluder}, and to the hidden persistent entity or region as the \emph{occludee}.
OCCUR-Bench contains 4,400 scenarios spanning 2- to 5-turn editing trajectories.
Each scenario consists of three functional stages: (i) occlusion, (ii) intermediate editing, and (iii) revelation.

In the occlusion stage, the occluder visually covers an occludee \(o_i\), making it temporarily invisible.
The occludee remains persistent, \(o_i \in \mathcal{P}_t\), because it is not semantically modified before revelation.
The intermediate stage may alter other regions or the global appearance, but leaves the occludee semantically unchanged.

A key design principle is that user instructions do not explicitly mention the occludee or directly request its restoration.
For example, an instruction may ask the editor to add an object in a plausible location or later remove it, without stating that the previously hidden content should be restored.
Thus, successful restoration requires preserving implicit scene content rather than merely following an explicit restoration command.

For each scenario, we define a restoration reference state:
\begin{equation}
I_{\mathrm{ref}}
\in
\mathcal{H}_{t-1},
\label{eq:reference_state}
\end{equation}
where \(I_{\mathrm{ref}}\) denotes the latest historical state in which the occludee is visible and semantically valid.
It provides the visual target for restoration, allowing OCCUR-Bench to distinguish faithful temporal restoration from plausible hallucination.

\subsection{Construction and Verification}
\label{subsec:construction_verification}

We summarize the construction procedure here and provide detailed scenario templates, dataset statistics, source image filtering criteria, and verification procedures in Appendix~\ref{app:occur_construction}.

We construct OCCUR-Bench from COCO~\cite{lin2014microsoft}, PIE-Bench~\cite{Ju2024_PIEBench}, and HQ-Edit~\cite{hui2024hq} images that contain salient occludee candidates, such as distinctive object parts, textures, logos, or clothing details.
Source images are selected only when a plausible occluder can be introduced without changing the viewpoint, object layout, or scene composition.
This filtering ensures that the benchmark primarily measures temporal preservation rather than failures caused by ill-posed editing instructions.

The benchmark covers diverse occlusion-and-revelation patterns built from common multi-turn editing operations.
Occlusion is typically induced by adding or transforming an occluder, while revelation is instantiated through operations such as removing, moving, resizing, or replacing the occluder.
Intermediate turns may include additional object insertions, attribute changes, or global style transformations, as long as they do not semantically modify the occludee.

All scenarios are manually verified before inclusion.
We remove cases with ambiguous occludees, insignificant hidden regions, insufficient occlusion, physically implausible occluder placement, or unnatural editing trajectories.
For longer sequences, we additionally verify that intermediate edits do not directly modify the occludee, ensuring that the final revelation turn evaluates temporal preservation rather than ordinary object editing.

\subsection{Evaluation Metrics}
\label{subsec:evaluation_metrics}
\begin{figure}[t]
    \centering
    \includegraphics[width=\linewidth]{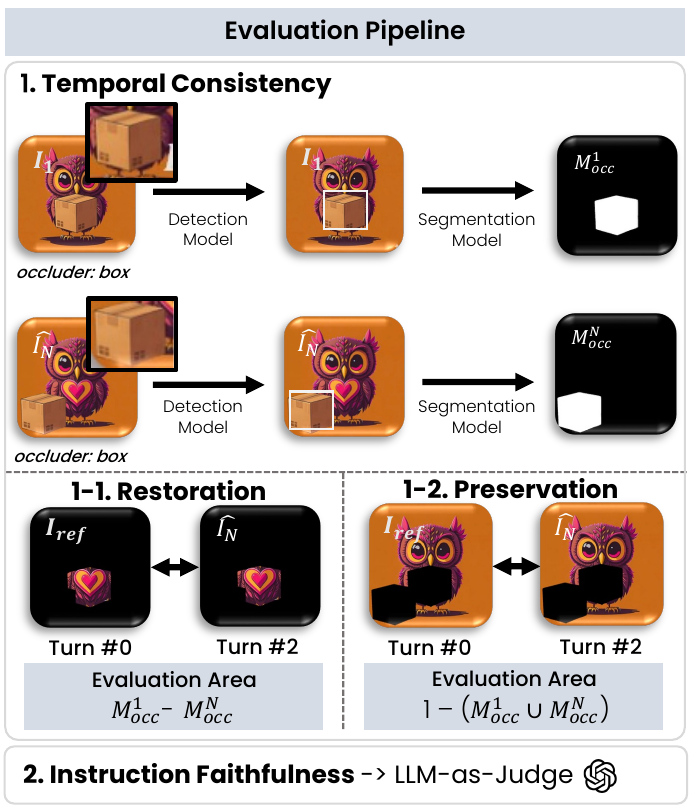}
\caption{OCCUR-Bench evaluation pipeline for temporal consistency.}    \label{fig:2_evaluation_pipeline}
    \vspace{-1em}
\end{figure}

As shown in Figure~\ref{fig:2_evaluation_pipeline}, OCCUR-Bench evaluates temporal visual consistency along two axes:
(i) restoration consistency and (ii) preservation consistency.
Restoration consistency measures whether temporally occluded content is faithfully recovered when it becomes visible again, while preservation consistency measures whether unchanged visible regions remain stable across the editing trajectory.
We additionally report instruction faithfulness as an auxiliary metric to verify that models execute the requested edit.

\noindent\underline{\textbf{Restoration Consistency.}}
Let \(M_{\mathrm{occ}}^{1:N}\) denote the set of pixels covered by the target occluder across the trajectory, and let \(M_{\mathrm{new}}^N\) denote the set of pixels occupied by newly introduced objects in the final revelation turn.
We define \(\mathcal{R}_{\mathrm{restore}}\subseteq\Omega\) as the pixel index set corresponding to the historically occluded region that becomes visible again after excluding final occluder and new-object regions, as illustrated in Figure~\ref{fig:2_evaluation_pipeline}.
Given the generated final image \(\hat{I}_N\) and the historical reference state \(I_{\mathrm{ref}}\), restoration consistency is defined as:
\begin{equation}
\mathcal{S}_{\mathrm{restore}}
=
\mathrm{sim}
\Big(
\hat{I}_N[\mathcal{R}_{\mathrm{restore}}],
I_{\mathrm{ref}}[\mathcal{R}_{\mathrm{restore}}]
\Big),
\label{eq:restoration_consistency}
\end{equation}
where \(I[\mathcal{R}]\) denotes the masked pixel subset indexed by \(\mathcal{R}\), and \(\mathrm{sim}(\cdot)\) computes the similarity between the generated and reference regions using the averaged normalized PSNR, LPIPS, and CLIP scores.

\noindent\underline{\textbf{Preservation Consistency.}}
Let \(M_{\mathrm{new}}^{1:N}\) denote the regions occupied by newly introduced objects across the trajectory.
We define \(\mathcal{R}_{\mathrm{preserve}}\) as the complement of \(M_{\mathrm{occ}}^{1:N}\) and \(M_{\mathrm{new}}^{1:N}\), corresponding to unchanged regions not involved in target occlusion or new object insertion.
Since \(I_{\mathrm{ref}}\) is a valid historical state for the unchanged scene content, we use it as the reference for preservation as well:
\begin{equation}
\mathcal{S}_{\mathrm{preserve}}
=
\mathrm{sim}
\Big(
\hat{I}_N[\mathcal{R}_{\mathrm{preserve}}],
I_{\mathrm{ref}}[\mathcal{R}_{\mathrm{preserve}}]
\Big).
\label{eq:preservation_consistency}
\end{equation}
For both metrics, \(\mathrm{sim}(\cdot)\) averages normalized PSNR, LPIPS-based similarity, and CLIP similarity.
Detailed construction of the masks and evaluation regions is provided in Appendix~\ref{sec:metric_details}.

\noindent\underline{\textbf{OCCUR-Bench Score.}}
The overall temporal consistency score is computed by averaging restoration and preservation consistency:

\begin{equation}
\mathcal{S}_{\mathrm{TC}}
=
\frac{1}{2}
\left(
\mathcal{S}_{\mathrm{restore}}
+
\mathcal{S}_{\mathrm{preserve}}
\right).
\label{eq:occur_score}
\end{equation}

This score separately accounts for whether a model restores temporally occluded content and preserves unchanged regions.

\noindent\underline{\textbf{Auxiliary Instruction Faithfulness.}}
We report instruction faithfulness \(\mathcal{S}_{\mathrm{IF}}\) using an LLM-as-a-judge protocol.
This score verifies whether the requested edit is executed and is not included in \(\mathcal{S}_{\mathrm{TC}}\).
\section{Proposed Framework}
\label{sec:method}

\begin{figure*}[t]
    \centering
   \includegraphics[width=\textwidth]{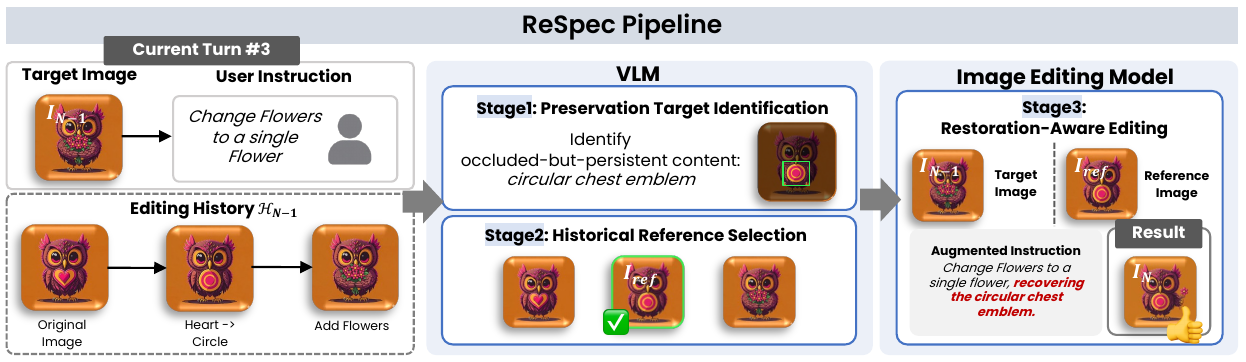} 
    \caption{
Overview of ReSpec.
A VLM-based controller infers the implicit preservation target from the editing history, selects a historical reference image that provides valid visual evidence, and rewrites the current instruction into a restoration-aware form.
The in-context editor then uses the current image, rewritten instruction, and selected reference to follow the requested edit while restoring temporally hidden content.
}
    \label{fig:4_resee}
    \vspace*{-0.5em}
\end{figure*}

We propose \textbf{\textit{ReSpec}} 
(\textbf{Re}ference \textbf{S}election and Preservation \textbf{Spec}ification), a training-free intent-grounded framework for conversational image editing.
Our key observation is that user instructions explicitly specify what should change, but often leave what should remain unchanged implicit.
This underspecification becomes problematic when unchanged content is no longer visible in the current image: the editor receives neither an explicit preservation request nor visual evidence for reconstruction.
ReSpec addresses this by using a VLM-based controller to make implicit preservation intent explicit.
The controller infers what content should persist, grounds this intent in a historical reference image, and realizes it as a restoration-aware instruction for the editor.
The framework proceeds in three steps:
(i) implicit preservation target inference,
(ii) historical reference grounding,
and
(iii) restoration-aware instruction realization.

\subsection{Preservation Target Identification}
\label{subsec:preservation_target_identification}

At turn \(t\), the VLM-based controller infers the preservation intent that is implicit in the editing history and current instruction.
It analyzes the instruction history, intermediate image states, and \(P_t\) to identify content that is not semantically modified but should remain recoverable.
This step produces \(D_t\), a textual description of the inferred preservation target.
For restoration turns, \(D_t\) identifies the occludee that should reappear after the occluder is removed or modified.
For non-restoration turns, \(D_t\) may be empty or describe visible content that should remain stable.
\subsection{Historical Reference Selection}
\label{subsec:historical_reference_selection}

When restoration is needed, the VLM-based controller grounds the inferred preservation target in historical visual evidence.
It selects \(I_{\mathrm{ref}}\) as the latest prior image in which the target is visible and semantically valid:
\begin{equation}
I_{\mathrm{ref}}
=
\operatorname{SelectRef}_{\mathrm{VLM}}(\mathcal{H}_{t-1}, P_t, D_t).
\label{eq:reference_selection}
\end{equation}
This reference supplies missing visual evidence for restoration; if no hidden content must be restored, ReSpec proceeds without reference selection.
\subsection{Restoration-Aware Editing}
\label{subsec:restoration_aware_editing}

Finally, the VLM-based controller realizes the inferred preservation intent as a restoration-aware instruction \(\tilde{P}_t\).
This instruction preserves the requested edit in \(P_t\) while explicitly stating the preservation target \(D_t\):
\begin{equation}
\tilde{P}_t
=
\operatorname{Rewrite}_{\mathrm{VLM}}(P_t, D_t).
\label{eq:instruction_rewrite}
\end{equation}
The in-context editor then generates the final image using the current image, rewritten instruction, and selected historical reference:
\begin{equation}
\hat{I}_t
=
f_{\mathrm{edit}}
(
I_{t-1},
\tilde{P}_t,
I_{\mathrm{ref}}
).
\label{eq:final_generation}
\end{equation}
For non-restoration turns, the editor can be applied without \(I_{\mathrm{ref}}\).
This procedure requires no additional training or architectural modification of the base editor; it only changes the textual and visual context provided at inference time.
\section{Experiments}
\label{sec:experiments}
We conduct experiments on OCCUR-Bench to answer the following three research questions:
\textbf{RQ1.}  How well do existing editors preserve temporally occluded content, and to what extent does ReSpec improve reference-conditioned editors?
\textbf{RQ2.} How do explicit preservation targets and historical reference grounding contribute to ReSpec?
\textbf{RQ3.} How does restoration consistency change as editing trajectories become longer?

\subsection{Experimental Setup}
\label{subsec:experimental_setup}

\noindent\underline{\textbf{Benchmark.}}
We evaluate conversational image editing systems on OCCUR-Bench, which contains 4,400 occlusion-and-revelation scenarios spanning 2- to 5-turn editing trajectories.
Each scenario provides a restoration reference \(I_{\mathrm{ref}}\), defined as the historical image state that contains valid visual evidence for the occludee.

\noindent\underline{\textbf{Baselines.}}
We use MC-Edit~\cite{Zhou2025_MultiTurnConsistentEditing} and Layer-wise Memory~\cite{Kim2025_ImprovingMultiTurn} as multi-turn baselines, evaluating their original outputs without applying our framework.
For reference-conditioned in-context editors, we evaluate Flux.2~\cite{blackforestlabs_flux2_2025} and OmniGen2~\cite{wu2025omnigen2} before and after applying our framework, since these models can accept selected historical references as input.
For Gemini-2.5, whose closed interface does not support controllable historical reference conditioning, we apply only the explicit-preservation component of our framework, which rewrites the instruction without supplying a historical reference image.

\noindent\underline{\textbf{Metrics.}}
We report restoration consistency \(\mathcal{S}_{\mathrm{restore}}\), preservation consistency \(\mathcal{S}_{\mathrm{preserve}}\), and their average temporal consistency score \(\mathcal{S}_{\mathrm{TC}}\).
We additionally report instruction faithfulness \(\mathcal{S}_{\mathrm{IF}}\) as an auxiliary metric to verify that temporal preservation improvements do not come at the cost of following the current instruction.

\noindent\underline{\textbf{Implementation Details.}}
We implement ReSpec as a two-stage inference pipeline: Qwen3-VL-8B-Instruct plans the preservation target, historical reference, and rewritten instruction in JSON format, and each target editor performs image synthesis.
For reference-conditioned editors, we provide the selected historical reference together with the current image; for Gemini-2.5, we use \texttt{gemini-2.5-flash-image} with only the rewritten instruction and current image, since controllable reference conditioning is unavailable.
We run \texttt{FLUX.2-klein-base-9B} with 30 denoising steps and guidance scale 4.0, and OmniGen2 with 50 denoising steps, text guidance 5.0, and image guidance 2.0.
We compute instruction faithfulness \(\mathcal{S}_{\mathrm{IF}}\) using GPT-4o-mini as an LLM-as-a-judge, and generate all outputs with fixed seeds when supported.

\begin{table}[t]
\centering
\small
\setlength{\tabcolsep}{5pt}
\resizebox{\columnwidth}{!}{
\begin{tabular}{lccc|c}
\toprule
Editor & $S_{\mathrm{restore}}$ & $S_{\mathrm{preserve}}$ & $S_{\mathrm{TC}}$ & $S_{\mathrm{IF}}$ \\
\midrule
MC-Edit & 0.323 & 0.367 & 0.345 & 0.271  \\
Layer-wise Memory & 0.291 & 0.290 & 0.290 & 0.273 \\
\midrule
Gemini-2.5 & 0.538 &  0.544 & 0.541 & 0.605 \\
Gemini-2.5 w/ ReSpec$^{\dagger}$ & \textbf{0.616} & \textbf{0.580} & \textbf{0.598} & \textbf{0.766}
 \\
\midrule
Flux.2 & 0.386 & 0.525 & 0.455 & 0.652 \\
Flux.2 w/ ReSpec & \textbf{0.547} & \textbf{0.620} & \textbf{0.584} & \textbf{0.756} \\
\midrule
OmniGen2 & 0.320 & 0.420 & 0.370 & 0.561 \\
OmniGen2 w/ ReSpec & \textbf{0.381} & \textbf{0.423} & \textbf{0.402} &  \textbf{0.691} \\
\bottomrule
\end{tabular}
}
\vspace{-0.5em}
\caption{
Main results on OCCUR-Bench.
We compare in-context editors with and without ReSpec.
\(\dagger\) denotes Gemini-2.5 without Historical Reference Selection due to interface constraints; Gemini variants use a 200-sample subset due to API cost.
}
\label{tab:1_main_results}
\vspace{-0.5em}

\end{table}

\begin{figure}[t]
    \centering
    \includegraphics[width=0.95\columnwidth]{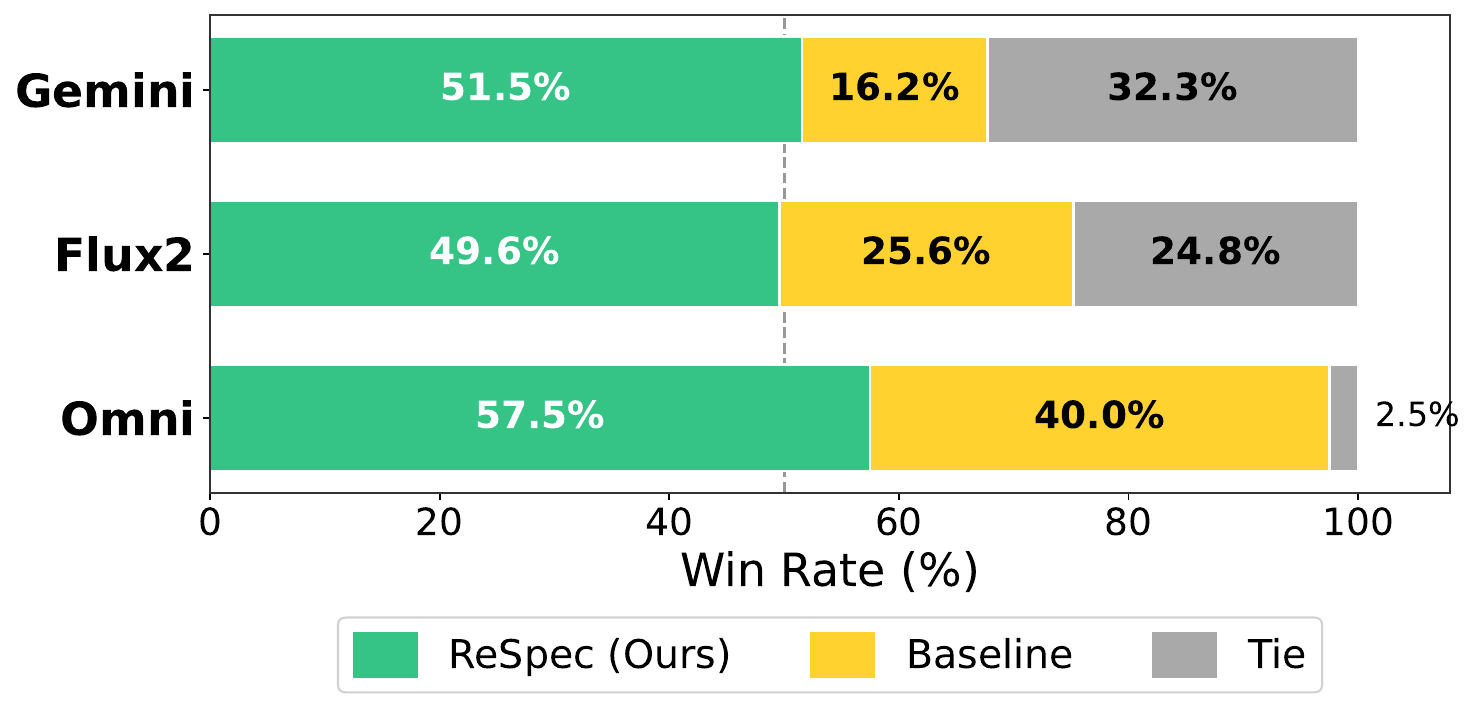}
\caption{
Human preference results for ReSpec versus each base editor.
Gemini and Flux.2 use response-scenario preferences, while OmniGen2 uses scenario-level mean ratings.
}
    \label{fig:human_preference}
    \vspace{-0.5em}

\end{figure}
\begin{table}[t]
\centering
\scriptsize
\setlength{\tabcolsep}{5pt}
\resizebox{\columnwidth}{!}{
\begin{tabular}{lccc|c}
\toprule
Editor & $S_{\mathrm{restore}}$ & $S_{\mathrm{preserve}}$ & $S_{\mathrm{TC}}$ & $S_{\mathrm{IF}}$ \\
\midrule
Flux.2 & 0.435 &  0.533 &  0.484 & 0.661 \\
Flux.2 + \textit{HRS}  & 0.480 & 0.552 &  0.517 & 0.647 \\
Flux.2 + \textit{PTI} & 0.540 & 0.558 & 0.549 & 0.687 \\
Flux.2 + \textit{PTI} + \textit{HRS} & \textbf{0.598} & \textbf{0.629} & \textbf{0.613} & \textbf{0.573}\\
\midrule
OmniGen2 & 	0.350 & 0.409 & 0.379 & 0.595 \\
OmniGen2 + \textit{HRS} & 0.358 & 0.420 & 0.389 & 0.606 \\
OmniGen2 + \textit{PTI} & 0.370 & 0.410 & 0.390 & 0.640\\
OmniGen2 + \textit{PTI} + \textit{HRS} & \textbf{0.427} & \textbf{0.427} & \textbf{0.427} & \textbf{0.701}  \\
\bottomrule
\end{tabular}
}
\vspace{-0.5em}
\caption{
Ablation study on Flux.2 and OmniGen2.
\textit{PTI} and \textit{HRS} are components of ReSpec, denoting Preservation Target Identification and Historical Reference Selection, respectively.
}
\label{tab:2_ablation_test}
\vspace{-1em}
\end{table}

\subsection{Experimental Results}
\label{subsec:experimental_results}

\noindent\underline{\textbf{Main Results.}}
Table~\ref{tab:1_main_results} answers RQ1 with two main findings.
First, existing conversational image editing models still struggle with temporal preservation under occlusion.
The multi-turn baselines achieve low temporal consistency scores, with MC-Edit and Layer-wise Memory reaching only 0.345 and 0.290 in \(S_{\mathrm{TC}}\), respectively.
This suggests that baseline models mainly preserve visible or spatially localized regions, but do not model content that is temporarily absent yet semantically persistent.

Second, ReSpec improves temporal preservation when applied to reference-conditioned in-context editors.
The improvement is most pronounced for Flux.2, where ReSpec increases \(S_{\mathrm{TC}}\) by \(+\)0.129, mainly through a \(+\)0.161 gain in restoration consistency.
For OmniGen2, ReSpec also improves \(S_{\mathrm{TC}}\), although with a smaller gain.
For Gemini-2.5, which does not support controllable historical reference conditioning, the explicit-preservation-only variant improves \(S_{\mathrm{TC}}\) by \(+\)0.057.

\noindent\underline{\textbf{Human Evaluation.}}
We conduct human evaluation to assess whether OCCUR-Bench scores align with human perception of temporal preservation.
We sample 40 scenarios and evaluate 8 model settings, resulting in 320 scenario-model outputs.
In total, we collect 960 raw human ratings from 12 evaluators, with each scenario-model output rated by exactly 3 evaluators.

Figure~\ref{fig:human_preference} shows that human evaluators generally favor ReSpec over the corresponding base editors, with preferred-or-tied rates of 83.85\% for Gemini and 74.40\% for Flux.2, and a 57.50\% scenario-level win rate for OmniGen2.
These results suggest that OCCUR-Bench improvements reflect human-perceived gains in restoration and preservation.
Automatic scores show moderate sample-level correlation with human ratings 
(\(r=0.409\) for \(S_{\mathrm{TC}}\)) and stronger model-level alignment 
(Pearson \(r=0.912\), Spearman \(\rho=0.881\), Kendall \(\tau=0.714\)), 
suggesting that OCCUR-Bench is especially reliable for aggregate model comparison.

\noindent\underline{\textbf{Ablation Test.}}
To answer RQ2, we conduct an ablation study on a 200-sample subset balanced by trajectory length and scenario type, ablating preservation target specification (\textit{PTI}) and historical reference selection (\textit{HRS}).
Table~\ref{tab:2_ablation_test} shows that both components improve temporal preservation and are complementary.
On Flux.2, \textit{PTI} provides a larger standalone gain, indicating that explicitly specifying the hidden preservation target is important for guiding restoration, while \textit{HRS} also improves \(S_{\mathrm{TC}}\) by grounding the edit in historical visual evidence.
Combining both components achieves the strongest performance, improving \(S_{\mathrm{TC}}\) by \(+\)0.129 and \(S_{\mathrm{restore}}\) by \(+\)0.163 over the Flux.2 baseline.
A similar trend holds for OmniGen2, where combining \textit{PTI} and \textit{HRS} also yields the strongest performance, improving \(S_{\mathrm{TC}}\) by \(+\)0.043 and \(S_{\mathrm{restore}}\) by \(+\)0.077 over the OmniGen2 baseline.
Overall, \textit{PTI} specifies what should be restored, whereas \textit{HRS} provides the visual evidence needed for faithful restoration.

\begin{figure*}[t]
\centering
\includegraphics[width=0.95\textwidth]{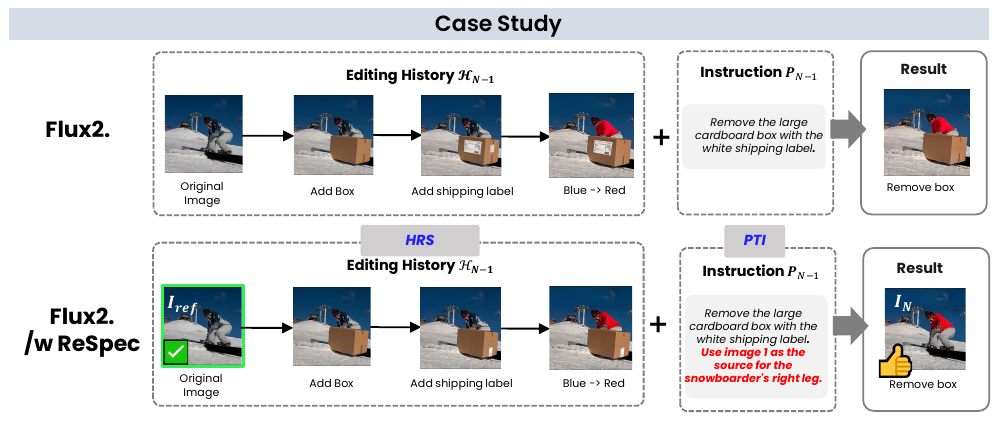}
\vspace{-0.5em}
\caption{
Qualitative case study of historical reference selection, where ReSpec restores an occluded target from an earlier source image while using the latest image as the editing anchor.
}
\label{fig:qualitative_case_study}
\vspace{-0.5em}
\end{figure*}
\begin{figure}[t]
    \centering
    \includegraphics[width=0.95\columnwidth]{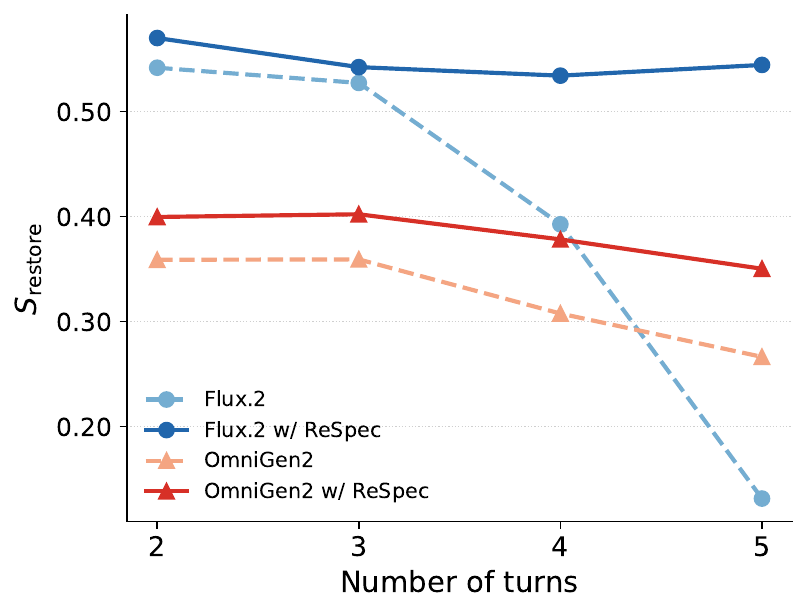}
\caption{
Restoration consistency across editing trajectory lengths on OCCUR-Bench.
Base editors degrade as trajectories become longer, while ReSpec maintains more stable restoration performance.
}
    \label{fig:5_turn_length_analysis}
    \vspace{-0.5em}

\end{figure}

\noindent\underline{\textbf{Case Study.}}
Figure~\ref{fig:qualitative_case_study} shows how ReSpec handles a restoration turn by identifying the snowboarder's right leg as the hidden preservation target and selecting the earlier image where it remains visible as the restoration source.
Using this historical reference with the latest image as the scene anchor enables the editor to recover the occluded target rather than hallucinating the revealed region.

\noindent\underline{\textbf{Analysis by Trajectory Length.}}
To answer RQ3, Figure~\ref{fig:5_turn_length_analysis} shows how restoration consistency changes as editing trajectories become longer.
We focus on \(\mathcal{S}_{\mathrm{restore}}\), since longer histories primarily make it harder to recover content that disappeared in earlier turns.
Flux.2 shows a sharp degradation, dropping from 0.542 at 2 turns to 0.131 at 5 turns, while ReSpec mitigates this collapse and maintains 0.544 at 5 turns.
OmniGen2 shows a more gradual decline from 0.359 to 0.266, and ReSpec consistently improves restoration performance across all trajectory lengths.
These results show that restoration consistency degrades as the needed content remains absent from the current image for longer histories.
ReSpec mitigates this long-horizon degradation by selecting a historical reference that provides explicit visual evidence for restoration.

\noindent\underline{\textbf{Additional Analyses.}}
Appendix~\ref{app:additional_analyses} provides further analyses of VLM-based controller quality, metric sensitivity, and computational cost.
We find that ReSpec is sensitive to the controller's ability to select the correct historical reference, since this selection determines whether the editor receives valid visual evidence for restoration (Appendix~\ref{app:reference_selection_quality}).
We also show that masked evaluation and restoration-region size affect automatic scores (Appendix~\ref{app:restoration_area_size}), with masked metrics better capturing localized restoration failures than whole-image similarity (Appendix~\ref{app:metric_analysis}).
Finally, we report the additional inference cost introduced by VLM-based control and reference-conditioned synthesis (Appendix~\ref{app:runtime_cost}).

\section{Related Work}
\label{sec:related_work}

Instruction-guided image editing has evolved from single-turn transformations~\cite{Brooks2023_InstructPix2Pix, cao2023masactrl} to conversational settings with iterative user instructions~\cite{Zhang2023_MagicBrush, Ge2024_SEEDDataEdit, Ye2025_ImgEdit}.
Existing methods improve multi-turn consistency through image references~\cite{Cui2023_ChatEdit, Zhou2025_MultiTurnConsistentEditing}, editing memory~\cite{Kim2025_ImprovingMultiTurn}, dialogue context~\cite{Ma2025_DialogDraw}, or agentic planning~\cite{Shen2026_IMAGAgent, Ye2026_AgentBanana, Gupta2025_CoSTA, Liang2025_RefineEditAgent, Yao2026_PhotoAgent}.
Meanwhile, benchmarks evaluate instruction following, final-output quality, content memory, backtracking, long-horizon task success~\cite{Zhang2023_MagicBrush, Ge2024_SEEDDataEdit, Ye2025_ImgEdit, Shen2026_IMAGAgent, Ye2026_AgentBanana, Liang2025_RefineEditAgent}, or region-aware editing quality~\cite{Ju2024_PIEBench, Li2025_BPM}.
However, both methods and evaluations primarily focus on visible or final-image content, leaving restoration of temporarily occluded content underexplored.
\section{Conclusion}
\label{sec:conclusion}

We introduced \textbf{\textit{OCCUR-Bench}}, a diagnostic benchmark for temporal preservation in conversational image editing under occlusion and revelation.
OCCUR-Bench shows that existing systems struggle to restore occluded-but-unchanged content.
We further proposed ReSpec, a training-free framework that makes implicit preservation explicit using restoration-aware instructions and historical visual references.
Experiments show that ReSpec improves restoration fidelity and temporal consistency, especially over longer trajectories.
These results motivate history-grounded temporal preservation beyond current-image preservation.

\section*{Limitations}

We acknowledge several limitations of the proposed benchmark and method.
First, our automatic evaluation depends on object detection and segmentation for constructing restoration and preservation regions.
Although masked evaluation is more sensitive to localized restoration failures than whole-image similarity, errors in open-vocabulary detection or mask extraction can still affect the resulting scores.
This limitation suggests the need for more robust region localization and human-aligned evaluation protocols.

Second, ReSpec depends on the quality of the VLM-based controller.
If the controller selects an incorrect historical reference or misidentifies the region that should be restored, the editor may receive misleading visual evidence.
Our analysis shows that reference selection quality is important for restoration performance, suggesting that stronger visual history understanding could further improve ReSpec.

Finally, ReSpec introduces additional inference overhead because it invokes a VLM controller during multi-turn editing.
While the framework is training-free and does not modify the base editor, this added computation may be non-negligible for long editing sessions.
Future work could reduce this cost through lightweight controllers, cached visual memory, or selective invocation only when restoration is likely required.

\bibliography{custom}

@String(CVPR= {IEEE Conf. Comput. Vis. Pattern Recog.})

@String(ICCV= {Int. Conf. Comput. Vis.})

@String(ICLR = {Int. Conf. Learn. Represent.})

@String(AAAI = {AAAI})

@String(CVPR  = {CVPR})

@String(ICCV  = {ICCV})

@String(ICLR  = {ICLR})

@inproceedings{Cui2023_ChatEdit,
  title     = {{C}hat{E}dit: Towards Multi-turn Interactive Facial Image Editing via Dialogue},
  author    = {Cui, Xing and Li, Zekun and Li, Peipei and Hu, Yibo and Shi, Hailin and Cao, Chunshui and He, Zhaofeng},
  booktitle = {Proceedings of the 2023 Conference on Empirical Methods in Natural Language Processing (EMNLP)},
  pages     = {14567--14583},
  year      = {2023},
  doi       = {10.18653/v1/2023.emnlp-main.899},
  url       = {https://aclanthology.org/2023.emnlp-main.899/}
}

@inproceedings{cao2023masactrl,
  title     = {MasaCtrl: Tuning-Free Mutual Self-Attention Control for Consistent Image Synthesis and Editing},
  author    = {Cao, Mingdeng and Wang, Xintao and Qi, Zhongang and Shan, Ying and Qie, Xiaohu and Zheng, Yinqiang},
  booktitle = {Proceedings of the IEEE/CVF International Conference on Computer Vision (ICCV)},
  pages     = {22560--22570},
  year      = {2023}
}

@misc{blackforestlabs_flux2_2025,
  author       = {{Black Forest Labs}},
  title        = {{FLUX.2: Frontier Visual Intelligence}},
  year         = {2025},
  howpublished = {GitHub repository},
  url          = {https://github.com/black-forest-labs/flux2},
  note         = {Official inference repository for FLUX.2 models}
}

@inproceedings{Zhou2025_MultiTurnConsistentEditing,
  title     = {Multi-turn Consistent Image Editing},
  author    = {Zhou, Zijun and Deng, Yingying and He, Xiangyu and Dong, Weiming and Tang, Fan},
  booktitle = {Proceedings of the IEEE/CVF International Conference on Computer Vision (ICCV)},
  year      = {2025}
}

@inproceedings{Kim2025_ImprovingMultiTurn,
  title     = {Improving Editability in Image Generation with Layer-wise Memory},
  author    = {Kim, Daneul and Lee, Jaeah and Park, Jaesik},
  booktitle = {Proceedings of the IEEE/CVF Conference on Computer Vision and Pattern Recognition (CVPR)},
  pages     = {7889--7898},
  year      = {2025}
}

@inproceedings{Brooks2023_InstructPix2Pix,
  title     = {InstructPix2Pix: Learning to Follow Image Editing Instructions},
  author    = {Brooks, Tim and Holynski, Aleksander and Efros, Alexei A.},
  booktitle = {Proceedings of the IEEE/CVF Conference on Computer Vision and Pattern Recognition (CVPR)},
  pages     = {18392--18402},
  year      = {2023}
}

@inproceedings{Zhang2023_MagicBrush,
  title     = {MagicBrush: A Manually Annotated Dataset for Instruction-Guided Image Editing},
  author    = {Zhang, Kai and Mo, Lingbo and Chen, Wenhu and Sun, Huan and Su, Yu},
  booktitle = {Advances in Neural Information Processing Systems (NeurIPS)},
  volume    = {36},
  pages     = {31428--31449},
  year      = {2023}
}

@inproceedings{lin2014microsoft,
  title={Microsoft coco: Common objects in context},
  author={Lin, Tsung-Yi and Maire, Michael and Belongie, Serge and Hays, James and Perona, Pietro and Ramanan, Deva and Doll{\'a}r, Piotr and Zitnick, C Lawrence},
  booktitle={European conference on computer vision},
  pages={740--755},
  year={2014},
  organization={Springer}
}

@article{hui2024hq,
  title={Hq-edit: A high-quality dataset for instruction-based image editing},
  author={Hui, Mude and Yang, Siwei and Zhao, Bingchen and Shi, Yichun and Wang, Heng and Wang, Peng and Zhou, Yuyin and Xie, Cihang},
  journal={arXiv preprint arXiv:2404.09990},
  year={2024}
}

@article{wu2025omnigen2,
  title={OmniGen2: Exploration to Advanced Multimodal Generation},
  author={Wu, Chenyuan and Zheng, Pengfei and Yan, Ruiran and Xiao, Shitao and Luo, Xin and Wang, Yueze and Li, Wanli and Jiang, Xiyan and Liu, Yexin and Zhou, Junjie and others},
  journal={arXiv preprint arXiv:2506.18871},
  year={2025}
}

@misc{Ge2024_SEEDDataEdit,
  title         = {SEED-Data-Edit Technical Report: A Hybrid Dataset for Instructional Image Editing},
  author        = {Ge, Yuying and Zhao, Sijie and Li, Chen and Ge, Yixiao and Shan, Ying},
  year          = {2024},
  eprint        = {2405.04007},
  archivePrefix = {arXiv},
  primaryClass  = {cs.CV},
  url           = {https://arxiv.org/abs/2405.04007}
}

@misc{Ye2025_ImgEdit,
  title         = {ImgEdit: A Unified Image Editing Dataset and Benchmark},
  author        = {Ye, Yang and He, Xianyi and Li, Zongjian and Lin, Bin and Yuan, Shenghai and Yan, Zhiyuan and Hou, Bohan and Yuan, Li},
  year          = {2025},
  eprint        = {2505.20275},
  archivePrefix = {arXiv},
  primaryClass  = {cs.CV},
  url           = {https://arxiv.org/abs/2505.20275}
}

@inproceedings{Ma2025_DialogDraw,
  title     = {{D}ialog{D}raw: Image Generation and Editing System Based on Multi-Turn Dialogue},
  author    = {Ma, Shichao and Zhang, Xinfeng and Zhao, Zeng and Liu, Bai and Fan, Changjie and Hu, Zhipeng},
  booktitle = {Proceedings of the AAAI Conference on Artificial Intelligence},
  volume    = {39},
  number    = {23},
  pages     = {24795--24803},
  year      = {2025},
  doi       = {10.1609/aaai.v39i23.34661}
}

@misc{Shen2026_IMAGAgent,
  title         = {{IMAGAgent}: Orchestrating Multi-Turn Image Editing via Constraint-Aware Planning and Reflection},
  author        = {Shen, Fei and Xie, Chengyu and Wang, Lihong and Zhang, Zhanyi and Jiang, Xin and Du, Xiaoyu and Tang, Jinhui},
  year          = {2026},
  eprint        = {2603.29602},
  archivePrefix = {arXiv},
  primaryClass  = {cs.GR},
  url           = {https://arxiv.org/abs/2603.29602}
}

@misc{Ye2026_AgentBanana,
  title         = {Agent Banana: High-Fidelity Image Editing with Agentic Thinking and Tooling},
  author        = {Ye, Ruijie and Zhang, Jiayi and Liu, Zhuoxin and Zhu, Zihao and Yang, Siyuan and Li, Li and Fu, Tianfu and Dernoncourt, Franck and Zhao, Yue and Zhu, Jiacheng and Rossi, Ryan and Chai, Wenhao and Tu, Zhengzhong},
  year          = {2026},
  eprint        = {2602.09084},
  archivePrefix = {arXiv},
  primaryClass  = {cs.CV},
  url           = {https://arxiv.org/abs/2602.09084}
}

@misc{Gupta2025_CoSTA,
  title         = {{CoSTA}$\ast$: Cost-Sensitive Toolpath Agent for Multi-turn Image Editing},
  author        = {Gupta, Advait and Velaga, NandaKiran and Nguyen, Dang and Zhou, Tianyi},
  year          = {2025},
  eprint        = {2503.10613},
  archivePrefix = {arXiv},
  primaryClass  = {cs.CV},
  url           = {https://arxiv.org/abs/2503.10613}
}

@misc{Liang2025_RefineEditAgent,
  title         = {An {LLM}-{LVLM} Driven Agent for Iterative and Fine-Grained Image Editing},
  author        = {Liang, Zihan and Sun, Jiahao and Ma, Haoran},
  year          = {2025},
  eprint        = {2508.17435},
  archivePrefix = {arXiv},
  primaryClass  = {cs.CV},
  url           = {https://arxiv.org/abs/2508.17435}
}

@misc{Yao2026_PhotoAgent,
  title         = {PhotoAgent: Agentic Photo Editing with Exploratory Visual Aesthetic Planning},
  author        = {Yao, Mingde and You, Zhiyuan and Man, Tam-King and Wang, Menglu and Xue, Tianfan},
  year          = {2026},
  eprint        = {2602.22809},
  archivePrefix = {arXiv},
  primaryClass  = {cs.CV},
  url           = {https://arxiv.org/abs/2602.22809}
}

@inproceedings{Ju2024_PIEBench,
  title     = {{PnP} Inversion: Boosting Diffusion-based Editing with 3 Lines of Code},
  author    = {Ju, Xuan and Zeng, Ailing and Bian, Yuxuan and Liu, Shaoteng and Xu, Qiang},
  booktitle = {The Twelfth International Conference on Learning Representations (ICLR)},
  year      = {2024},
  url       = {https://openreview.net/forum?id=FoMZ4ljhVw}
}

@inproceedings{Li2025_BPM,
  title     = {Balancing Preservation and Modification: A Region and Semantic Aware Metric for Instruction-Based Image Editing},
  author    = {Li, Zhuoying and Xu, Zhu and Peng, Yuxin and Liu, Yang},
  booktitle = {Proceedings of the International Conference on Machine Learning (ICML)},
  year      = {2025},
  url       = {https://openreview.net/forum?id=Nrs6csi52N}
}

\clearpage
\appendix
\begin{center}
{\Large \textbf{Appendix}}
\end{center}
\vspace{1em}

\section{OCCUR-Bench Construction Details}
\label{app:occur_construction}

\subsection{Source Image Selection and Filtering}
\label{app:source_image_selection}

We construct OCCUR-Bench from source images collected from existing vision and image editing datasets, including COCO, PIE-Bench, and HQEdit.
Since OCCUR-Bench is designed to evaluate occlusion-and-revelation scenarios, we do not use all images from these datasets.
Instead, we manually curate images that contain visually identifiable content that can be temporarily occluded and later revealed.

During image selection, we prioritize images with at least one salient occludee candidate, such as a distinctive object part, texture, logo, body region, clothing detail, or background element.
The occludee should be visually meaningful and large enough to support reliable evaluation after revelation.
We also require that a plausible occluder can be inserted without changing the camera viewpoint, object layout, or global scene composition.

We exclude images where the potential occludee is too small, visually ambiguous, or difficult to evaluate.
We also remove images for which a plausible occlusion would require physically implausible placement, large pose changes, camera movement, or substantial modification of existing scene elements.
This filtering reduces ambiguity and ensures that OCCUR-Bench primarily measures temporal preservation failures rather than failures caused by ill-posed editing instructions.

\subsection{Scenario Generation and Human Verification}
\label{app:scenario_generation_verification}

Given a selected source image, we generate candidate multi-turn editing scenarios using a VLM-based scenario generation pipeline.
The generator proposes an occludee, an occluding object, turn-level editing instructions, and the restoration source state.
Each scenario follows the occlusion-and-revelation structure described in the main text: an occluder first hides a persistent entity, optional intermediate edits are applied without directly modifying that entity, and a later revelation edit exposes the hidden region again.

A key design principle is that the default user instructions do not explicitly mention the occludee or ask the model to restore it.
For example, an occlusion instruction may ask the editor to add an object in a plausible location, rather than explicitly saying that it should cover a specific target.
Similarly, the revelation instruction modifies, moves, shrinks, replaces, or removes the occluder without explicitly requesting restoration of the hidden content.
This preserves the implicit nature of temporal preservation: the hidden content should be recovered because it remains semantically persistent, not because the instruction directly asks for it.

All generated scenarios are manually verified before inclusion in OCCUR-Bench.
We remove scenarios with ambiguous occludees, insignificant hidden regions, insufficient occlusion, physically implausible occluder placement, or unnatural editing trajectories.
For longer sequences, we additionally check that intermediate edits do not directly modify the occludee.
This ensures that the final revelation turn evaluates temporal preservation rather than ordinary object editing.

\subsection{Scenario Templates and Dataset Statistics}
\label{app:scenario_templates}

OCCUR-Bench uses three functional edit stages: occlusion, intermediate editing, and revelation.
The occlusion stage is instantiated by object addition.
For revelation, we consider object replacement, object relocation, object removal, and object shrinking.
Intermediate edits include adding new objects, changing the color, material, or pattern of existing objects, and applying global style transformations.

\begin{table}[t]
\centering
\scriptsize
\setlength{\tabcolsep}{4pt}
\begin{tabular}{l l r}
\toprule
Length & Template & \#Scen. \\
\midrule
2-turn & $I_0 \rightarrow O(T_1) \rightarrow R(T_2)$ & 1,000 \\
3-turn & $I_0 \rightarrow O(T_1) \rightarrow M(T_2) \rightarrow R(T_3)$ & 1,000 \\
4-turn v1 & $I_0 \rightarrow O(T_1) \rightarrow M(T_2) \rightarrow M(T_3) \rightarrow R(T_4)$ & 1,000 \\
4-turn v2 & $I_0 \rightarrow S(T_1) \rightarrow O(T_2) \rightarrow M(T_3) \rightarrow R(T_4)$ & 200 \\
5-turn v1--v4 & $I_0 \rightarrow \cdots \rightarrow O(T_k) \rightarrow \cdots \rightarrow R(T_5)$, $k{\in}\{1,2,3,4\}$ & 1,000 \\
5-turn v5 & $I_0 \rightarrow S(T_1) \rightarrow O(T_2) \rightarrow M(T_3) \rightarrow M(T_4) \rightarrow R(T_5)$ & 200 \\
\midrule
Total & & 4,400 \\
\bottomrule
\end{tabular}
\caption{
Scenario templates and count distribution in OCCUR-Bench.
$O$, $M$, $S$, and $R$ denote occlusion, intermediate edit, style transformation, and revelation, respectively.
}
\label{tab:scenario_templates}
\end{table}

\subsection{Ground-Truth Historical Reference Annotation}
\label{app:gt_reference_annotation}

For each scenario, we annotate a ground-truth restoration reference state.
This reference is the historical image state that provides the correct visual target for the content that becomes visible again at the revelation turn.

The ground-truth reference is not necessarily the original image.
If the persistent entity was edited before being occluded, the correct reference is the edited state before occlusion, not the initial image.
In general, we define the ground-truth restoration reference as the latest historical image in which the persistent entity is visible and semantically valid with respect to the editing history.

This annotation is used for evaluation and oracle ablations only.
It is not provided to the editing models or to the ReSpec orchestrator during inference.
By separating the ground-truth reference from the model input, OCCUR-Bench evaluates whether a model can recover temporally persistent content from the available editing history rather than relying on explicit access to the answer.

\subsection{Details of Evaluation Metrics}
\label{sec:metric_details}

This section provides the detailed mask construction procedure used to compute restoration consistency and preservation consistency in OCCUR-Bench.

\paragraph{Mask Extraction.}
For each editing turn \(t\), we derive two types of binary masks using an open-vocabulary detection-and-segmentation pipeline.
The target-occluder mask \(M_{\mathrm{occ}}^t\) indicates the region occupied by the annotated occluder, while the new-object mask \(M_{\mathrm{new}}^t\) indicates the union of regions occupied by newly introduced objects.
If the annotated target occluder or newly introduced objects are absent at turn \(t\), the corresponding mask is set to the empty mask.
Let \(N\) denote the final revelation turn.

\paragraph{Restoration Region.}
To identify the region where temporally hidden content should reappear, we first aggregate all regions covered by the target occluder throughout the trajectory:
\begin{equation}
M_{\mathrm{occ}}^{1:N}
=
\bigcup_{t=1}^{N} M_{\mathrm{occ}}^t.
\label{eq:appendix_occ_union}
\end{equation}
The restoration region is then defined as the part of the historically occluded area that is no longer occupied by the target occluder or newly introduced objects in the final image:
\begin{equation}
\mathcal{R}_{\mathrm{restore}}
=
M_{\mathrm{occ}}^{1:N}
\setminus
\left(
M_{\mathrm{occ}}^N
\cup
M_{\mathrm{new}}^N
\right).
\label{eq:appendix_restore_region}
\end{equation}
This region approximates the area where the previously hidden occludee should become visible again after the occluder is removed, moved, resized, or replaced.

Given the generated final image \(\hat{I}_N\) and the valid historical reference state \(I_{\mathrm{ref}}\), restoration consistency is computed as:
\begin{equation}
\mathcal{S}_{\mathrm{restore}}
=
\mathrm{sim}
\Big(
\hat{I}_N(\mathcal{R}_{\mathrm{restore}}),
I_{\mathrm{ref}}(\mathcal{R}_{\mathrm{restore}})
\Big).
\label{eq:appendix_restoration_consistency}
\end{equation}
Here, \(\hat{I}_N(\mathcal{R}_{\mathrm{restore}})\) and \(I_{\mathrm{ref}}(\mathcal{R}_{\mathrm{restore}})\) denote the corresponding restoration regions in the generated final image and the historical reference state, respectively.

\paragraph{Preservation Region.}
Preservation consistency evaluates whether regions unrelated to the target occlusion or new object insertion remain visually stable.
Let \(\Omega\) denote the full image domain.
We first aggregate all regions occupied by newly introduced objects throughout the trajectory:
\begin{equation}
M_{\mathrm{new}}^{1:N}
=
\bigcup_{t=1}^{N} M_{\mathrm{new}}^t.
\label{eq:appendix_new_union}
\end{equation}
The preservation region is defined as the complement of all regions involved in target occlusion or new object insertion:
\begin{equation}
\mathcal{R}_{\mathrm{preserve}}
=
\Omega
\setminus
\left(
M_{\mathrm{occ}}^{1:N}
\cup
M_{\mathrm{new}}^{1:N}
\right).
\label{eq:appendix_preserve_region}
\end{equation}
This region corresponds to unchanged background or non-target content that should remain visually stable across the editing trajectory.

Preservation consistency is computed as:
\begin{equation}
\mathcal{S}_{\mathrm{preserve}}
=
\mathrm{sim}
\Big(
\hat{I}_N(\mathcal{R}_{\mathrm{preserve}}),
I_{\mathrm{ref}}(\mathcal{R}_{\mathrm{preserve}})
\Big).
\label{eq:appendix_preservation_consistency}
\end{equation}
Here, \(\hat{I}_N(\mathcal{R}_{\mathrm{preserve}})\) and \(I_{\mathrm{ref}}(\mathcal{R}_{\mathrm{preserve}})\) denote the corresponding unchanged regions in the generated final image and the historical reference state, respectively.

\paragraph{Similarity Function.}
For both restoration and preservation consistency, the similarity function 
\(\mathrm{sim}(\cdot)\) is computed as the average of normalized PSNR, LPIPS, and CLIP similarity:
\begin{equation}
\begin{aligned}
\mathrm{sim}(A,B)
= \frac{1}{3} \big(
& \mathrm{PSNR}_{\mathrm{norm}}(A,B) \\
& + \left(1 - \mathrm{LPIPS}_{\mathrm{norm}}(A,B)\right) \\
& + \mathrm{CLIP}(A,B)
\big).
\end{aligned}
\label{eq:appendix_similarity}
\end{equation}
All components are normalized to the same range before averaging, so higher values indicate greater visual consistency.

\paragraph{Temporal Consistency Score.}
The overall temporal consistency score is computed by averaging restoration and preservation consistency:
\begin{equation}
\mathcal{S}_{\mathrm{TC}}
=
\frac{1}{2}
\left(
\mathcal{S}_{\mathrm{restore}}
+
\mathcal{S}_{\mathrm{preserve}}
\right).
\label{eq:appendix_occur_score}
\end{equation}
This score evaluates whether a model both restores temporally occluded content and preserves unchanged visible content.

\paragraph{Instruction Faithfulness.}
In addition to temporal consistency, we report instruction faithfulness \(\mathcal{S}_{\mathrm{IF}}\) using an LLM-as-a-judge protocol.
This auxiliary score assesses whether the generated final image follows the current editing instruction.
It is not included in \(\mathcal{S}_{\mathrm{TC}}\), which is designed to specifically measure temporal visual consistency under occlusion and revelation.

\section{Evaluation Details}
\label{app:evaluation_details}

\subsection{Open-Vocabulary Detection and Mask Extraction}
\label{app:mask_extraction}

OCCUR-Bench provides semantic annotations rather than manually annotated segmentation masks.
For region-aware evaluation, we derive object masks using an open-vocabulary detection and segmentation pipeline.
We use YOLO-Worldv2 with the \texttt{yolov8l-worldv2.pt} checkpoint for object detection, followed by SAM with the \texttt{vit-h} checkpoint for box-conditioned mask extraction.

For each scenario, we construct detection queries from the annotated target occluder and newly introduced object labels.
We do not directly detect the occludee, since the restoration region is defined from the occluder region that becomes visible again after revelation.
YOLO-Worldv2 is run with a low internal confidence threshold of \texttt{conf=0.001} to avoid prematurely discarding candidate boxes.
We then apply a separate acceptance threshold of 0.2 to the returned detections.
When multiple boxes are returned for the same query, we use the highest-confidence box.

To improve robustness to phrasing differences in object labels, we use a simple query fallback chain.
We first use the original object label.
If no detection passes the acceptance threshold, we retry using the last word of the label, and then a simple singularized form.
We record the accepted prompt, fallback stage, confidence score, and bounding box in the detection output.

The selected bounding box is passed to SAM to obtain a binary mask.
Bounding boxes are converted to pixel coordinates when necessary and clipped to the image boundary.
If no detection passes the threshold, the bounding box is invalid, or SAM fails, we assign an all-zero black mask with the same resolution as the corresponding image.
We do not manually correct these masks.

\subsection{Human Evaluation Protocol}
\label{app:human_eval_protocol}

We conduct a human evaluation on 40 OCCUR-Bench scenarios balanced across sequence lengths.
For each scenario, annotators evaluate outputs from eight models, resulting in 320 unique model-scenario outputs.
Each model-scenario output is rated by three independent annotators, yielding 960 raw ratings.
Each annotator evaluates 10 scenarios and rates outputs from all eight models for each assigned scenario.

Annotators are shown the relevant editing history, the current instruction, and the final generated image.
They evaluate each output along three dimensions: restoration, preservation, and instruction following.
Restoration measures whether the temporally occluded target content is correctly recovered in the final image.
Preservation measures whether other unchanged regions remain visually consistent with the editing history.
Instruction following measures whether the final output correctly executes the current editing instruction.
Cases where the requested revelation edit is not executed are treated as unsuccessful.

For each output, we average the ratings from three annotators to obtain human restoration, preservation, and instruction-following scores.
We compare these human scores with the corresponding automatic metrics using Pearson, Spearman, and Kendall correlations.
For instruction following, we use the mean automatic instruction-following score rather than the sum, since the sum is biased by sequence length.

\section{ReSpec Implementation Details}
\label{app:ReSpec_implementation}

\subsection{Visual Intent Orchestrator Inputs and Outputs}
\label{app:orchestrator_io}

We implement ReSpec as a VLM-based visual intent orchestrator.
At turn \(t\), the orchestrator receives the full image history up to the current input image, \(\{I_0, I_1, \ldots, I_{t-1}\}\), together with the instruction history and the current user instruction.
Here, \(I_0\) is the original image and \(I_{t-1}\) is the current image provided to the editor.
We do not duplicate \(I_{t-1}\) as a separate input.

The first editing turn is executed without a VLM call.
For turn 1, the base editor directly receives the original image \(I_0\) and the first user instruction.
Starting from turn 2, ReSpec invokes the VLM orchestrator once per turn.

For restoration turns, the orchestrator returns a structured JSON object with the following fields:
\texttt{reasoning}, \texttt{identity\_anchor\_id}, \texttt{restoration\_source\_id}, \texttt{restoration\_region}, \texttt{image\_ids}, and \texttt{synthesis\_instruction}.
The \texttt{restoration\_source\_id} specifies the historical image used as visual evidence for restoration, while \texttt{identity\_anchor\_id} specifies the current or recent image used to preserve the active edited state.
The \texttt{image\_ids} field determines which images are passed to the editor, and \texttt{synthesis\_instruction} is used as the final editing instruction.

For non-restoration intermediate turns, the orchestrator only returns \texttt{reasoning} and \texttt{synthesis\_instruction}.
In these cases, ReSpec does not perform historical reference selection and the editor receives only the current image \(I_{t-1}\) along with the rewritten instruction.

Ground-truth reference ids are not provided to the VLM prompt or image input.
The runner uses ground-truth fields only to determine whether the final turn is a restoration turn and to store logging and evaluation metadata.
Thus, the VLM decisions are made from the visual history, instruction history, and current user request, without direct access to the ground-truth restoration source.

\subsection{Prompt Templates and Parsing Rules}
\label{app:prompt_parsing}

We use separate prompt templates for restoration turns and non-restoration intermediate turns.
For Qwen, these correspond to \texttt{qwen\_vpe\_orchestrator\_restoration} and \texttt{qwen\_vpe\_orchestrator\_midturn}.
InternVL uses the same prompt structure; when an InternVL-specific prompt key is unavailable, the runner falls back to the corresponding Qwen prompt key.

The restoration prompt asks the VLM to analyze the image and instruction history, determine whether the current instruction reveals previously occluded content, select the appropriate historical restoration source, select an identity anchor image, and produce a concise synthesis instruction for the editor.
The mid-turn prompt asks the VLM to rewrite the current user request into a synthesis instruction while preserving the already established visual state.

The VLM is instructed to return a strict JSON object.
During parsing, the runner first searches for a fenced \texttt{json} block.
If no such block is found, it extracts the first JSON-like object using a regular expression.
The parsed JSON fields are then converted into the editor inputs.

If JSON parsing fails, the runner constructs a fallback object.
For restoration turns, the reference ids are returned as \texttt{None} and \texttt{image\_ids} is returned as an empty list, while the raw VLM response is stored as \texttt{synthesis\_instruction}.
For mid-turns, the raw VLM response is used as \texttt{synthesis\_instruction}.
If \texttt{synthesis\_instruction} is missing or the VLM result is not a valid dictionary, the current user request is used as the fallback instruction.

The final \texttt{synthesis\_instruction} is passed directly to the editor.
It is not merged with the original user instruction after parsing.

\subsection{Fallback Rules for Invalid VLM Outputs}
\label{app:vlm_fallback}

We apply deterministic fallback rules when the VLM output is incomplete or invalid.
These rules do not use the ground-truth restoration source.

For restoration turns, if \texttt{identity\_anchor\_id} is missing or null, we set it to \(t-1\), corresponding to the current editor input image.
If \texttt{restoration\_source\_id} is missing or null, we set it to 0, corresponding to the original image.
If \texttt{restoration\_region} is empty, it is left as \texttt{None}.
If \texttt{image\_ids} is missing or empty, we reconstruct it as \([\texttt{restoration\_source\_id}, \texttt{identity\_anchor\_id}]\).

We do not clamp out-of-range image ids during VLM output parsing.
Instead, when ordering images for the editor, the runner keeps only ids that exist in the available image pool.
If all selected ids are invalid, the runner falls back to the latest available image, usually \(I_{t-1}\).

For non-restoration turns, ReSpec does not use historical reference selection.
The editor receives only the current image \(I_{t-1}\), and the prompt is the VLM-rewritten \texttt{synthesis\_instruction}.

\subsection{Runtime and Cost}
\label{app:runtime_cost}

ReSpec is training-free and does not modify the architecture or weights of the base editor.
The additional cost comes from the local VLM orchestration step used to analyze the editing history and rewrite the editing instruction.

In our implementation, both Qwen and InternVL are loaded as local models.
We use \texttt{Qwen3-VL-8B-Instruct} for Qwen and \texttt{InternVL3-8B-Instruct} for InternVL.
The ReSpec runner does not call GPT or Gemini APIs.

Under the unified runner, the first editing turn does not require a VLM call.
For every subsequent turn, including both restoration and non-restoration turns, the runner makes one local VLM call.
Thus, a \(T\)-turn scenario requires \(T-1\) VLM calls.
The base editor is still executed once per turn, as in the original editing pipeline.
Therefore, the additional runtime overhead of ReSpec comes from local VLM inference, while the editor-side inference cost remains unchanged.

\section{Additional Analyses}
\label{app:additional_analyses}
\subsection{Effect of Revelation Type}
\label{app:revelation_type}

We analyze performance across different revelation operations.
Revelation operations differ in how the previously occluded region becomes visible again, and therefore may pose different challenges for temporal preservation.
\begin{table}[t]
\centering
\scriptsize
\setlength{\tabcolsep}{5pt}
\resizebox{\linewidth}{!}{
\begin{tabular}{llcccc}
\toprule
Type & Description 
& Flux.2 & Flux.2w/ ReSpec
& OmniGen2 & OmniGen2 w/ ReSpec \\
\midrule
A & Resize
& 0.460 & 0.557 
& 0.370 & 0.354 \\
B & Move 
& 0.474 & 0.528 
& 0.399 & 0.403 \\
C & Replace
& 0.427 & 0.621 
& 0.364 & 0.410 \\
D & Remove
& 0.456 & 0.636 
& 0.358 & 0.439 \\
E & Multi-edit 
& 0.458 & 0.594 
& 0.354 & 0.427 \\
\bottomrule
\end{tabular}
}
\caption{
Temporal consistency by revelation type.
Scores are reported using \(S_{\mathrm{TC}}\).
}
\label{tab:revelation_type}
\end{table}

Table~\ref{tab:revelation_type} shows that ReSpec improves Flux.2 across all revelation types, with the largest gains for replacement \((+0.194)\), removal \((+0.180)\), and multi-occluder cases \((+0.136)\).
For OmniGen2, the gains are smaller and less uniform: ReSpec improves most revelation types, especially removal \((+0.081)\) and multi-occluder cases \((+0.073)\), but slightly decreases performance for resize cases \((-0.016)\).
Overall, these results suggest that historical reference grounding is most helpful when the final edit clearly reveals previously hidden content.

\subsection{Effect of Reference Distance}
\label{app:reference_distance}

We further analyze temporal consistency by reference distance, defined as the number of turns between the restoration reference and the final revelation turn.

\begin{table}[t]
\centering
\scriptsize
\setlength{\tabcolsep}{4pt}
\begin{tabular}{lcccc}
\toprule
Distance & Flux.2 & Flux.2 w/ ReSpec & OmniGen2 & OmniGen2 w/ ReSpec \\
\midrule
\(\Delta=2\) & 0.527 & 0.647 & 0.407 & 0.436 \\
\(\Delta=3\) & 0.452 & 0.576 & 0.368 & 0.395 \\
\(\Delta=4\) & 0.405 & 0.554 & 0.344 & 0.382 \\
\(\Delta=5\) & 0.411 & 0.482 & 0.350 & 0.385 \\
\bottomrule
\end{tabular}
\caption{
Temporal consistency by reference distance.
Scores are reported using \(S_{\mathrm{TC}}\).
}
\label{tab:reference_distance}
\end{table}

Table~\ref{tab:reference_distance} shows that temporal consistency generally decreases as the relevant historical reference becomes more distant.
ReSpec consistently improves both Flux.2 and OmniGen2 across all reference distances, indicating that historical reference grounding remains beneficial even when the needed visual evidence appears several turns earlier.
The gain is especially pronounced for Flux.2 at intermediate distances, where the base editor substantially degrades without historical visual grounding.

\subsection{Effect of Restoration Area Size}
\label{app:restoration_area_size}

We analyze whether the size of the restoration region affects temporal consistency.
We divide examples into three bins according to the restoration area ratio, defined as the area of the restoration region divided by the full image area.

\begin{table}[t]
\centering
\scriptsize
\setlength{\tabcolsep}{6pt}
\begin{tabular}{lccc}
\toprule
Area Bin & Flux.2 & Flux.2 w/ ReSpec & Gain \\
\midrule
Small  & 0.618 & 0.649 & +0.031 \\
Medium & 0.620 & 0.690 & +0.070 \\
Large  & 0.550 & 0.672 & +0.122 \\
\bottomrule
\end{tabular}
\caption{
Temporal consistency by restoration area size.
Scores are reported using \(S_{\mathrm{TC}}\).
Area bins are defined by tertiles of the restoration area ratio.
}
\label{tab:restoration_area_size}
\end{table}

Table~\ref{tab:restoration_area_size} shows that larger restoration regions are more challenging for the base editor.
Flux.2 achieves \(S_{\mathrm{TC}}=0.618\) in the small-area bin and drops to 0.550 in the large-area bin.
ReSpec improves temporal consistency across all area bins, with the largest gain in the large-area bin \((+0.122)\).
This suggests that historical reference grounding is especially useful when a larger hidden region must be restored.

\subsection{Historical Reference Selection Quality}
\label{app:reference_selection_quality}

We analyze historical reference selection quality on a shared 200-sample diagnostic subset.
This analysis compares Qwen and InternVL as VLM history analyzers using the same base editor and evaluation protocol.
Reference selection accuracy is defined as the fraction of examples where the predicted \texttt{restoration\_source\_id} matches the ground-truth historical reference.

\begin{table}[t]
\centering
\scriptsize
\setlength{\tabcolsep}{5pt}
\begin{tabular}{lccccc}
\toprule
Analyzer & Ref. Acc. & \(S_{\mathrm{restore}}\) & \(S_{\mathrm{preserve}}\) & \(S_{\mathrm{TC}}\) & \(N\) \\
\midrule
Qwen    & 79.0\% & 0.398 & 0.367 & 0.383 & 200 \\
InternVL & 59.0\% & 0.347 & 0.364 & 0.355 & 200 \\
\bottomrule
\end{tabular}
\caption{
Historical reference selection quality on a shared 200-sample diagnostic subset.
\(N\) denotes the number of common sessions used for both analyzers.
}
\label{tab:reference_selection_quality}
\end{table}

Table~\ref{tab:reference_selection_quality} shows that Qwen selects the correct historical reference more accurately than InternVL, improving reference selection accuracy by 20.0 percentage points.
This difference is reflected mainly in restoration consistency: Qwen improves \(S_{\mathrm{restore}}\) from 0.347 to 0.398, while \(S_{\mathrm{preserve}}\) remains nearly unchanged.
These results suggest that reference selection quality primarily affects the restoration region and supports the use of Qwen as the default history analyzer in ReSpec.

\subsection{Metric Analysis}
\label{app:metric_analysis}

\begin{figure}[t]
\centering
\includegraphics[width=\linewidth]{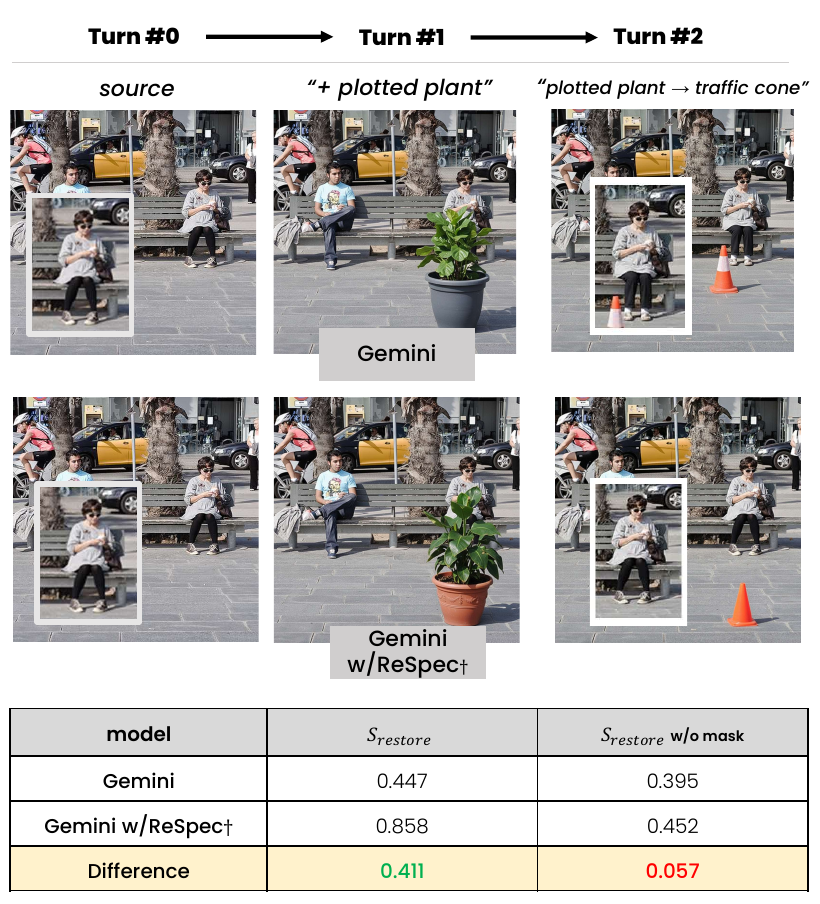}
\vspace{-0.5em}
\caption{
Qualitative example illustrating the limitation of whole-image similarity for restoration evaluation.
Although the requested edit is applied, the restoration failure is localized to the small revealed region, which is better captured by masked restoration scoring than by whole-image metrics.
}
\label{fig:metric_study}
\vspace{-0.5em}
\end{figure}

Figure~\ref{fig:metric_study} illustrates why OCCUR-Bench uses masked restoration scoring rather than whole-image similarity.
Although both models replace the potted plant with a traffic cone, the key difference appears only in the small revealed region where the person should be restored.
Whole-image CLIP, LPIPS, and PSNR are dominated by unchanged background areas and therefore show only a small score difference, while the masked restoration score clearly captures the restoration failure.

\section{Prompt Templates}
\label{app:prompt_templates}

This section provides the original prompt templates used in our benchmark construction, instruction-following evaluation, and ReSpec orchestration.
Specifically, we include prompts for scenario generation, scenario extension, instruction-following evaluation, restoration-aware orchestration, and non-restoration instruction rewriting.

\begin{figure*}[t]
\centering
\includegraphics[width=0.95\textwidth]{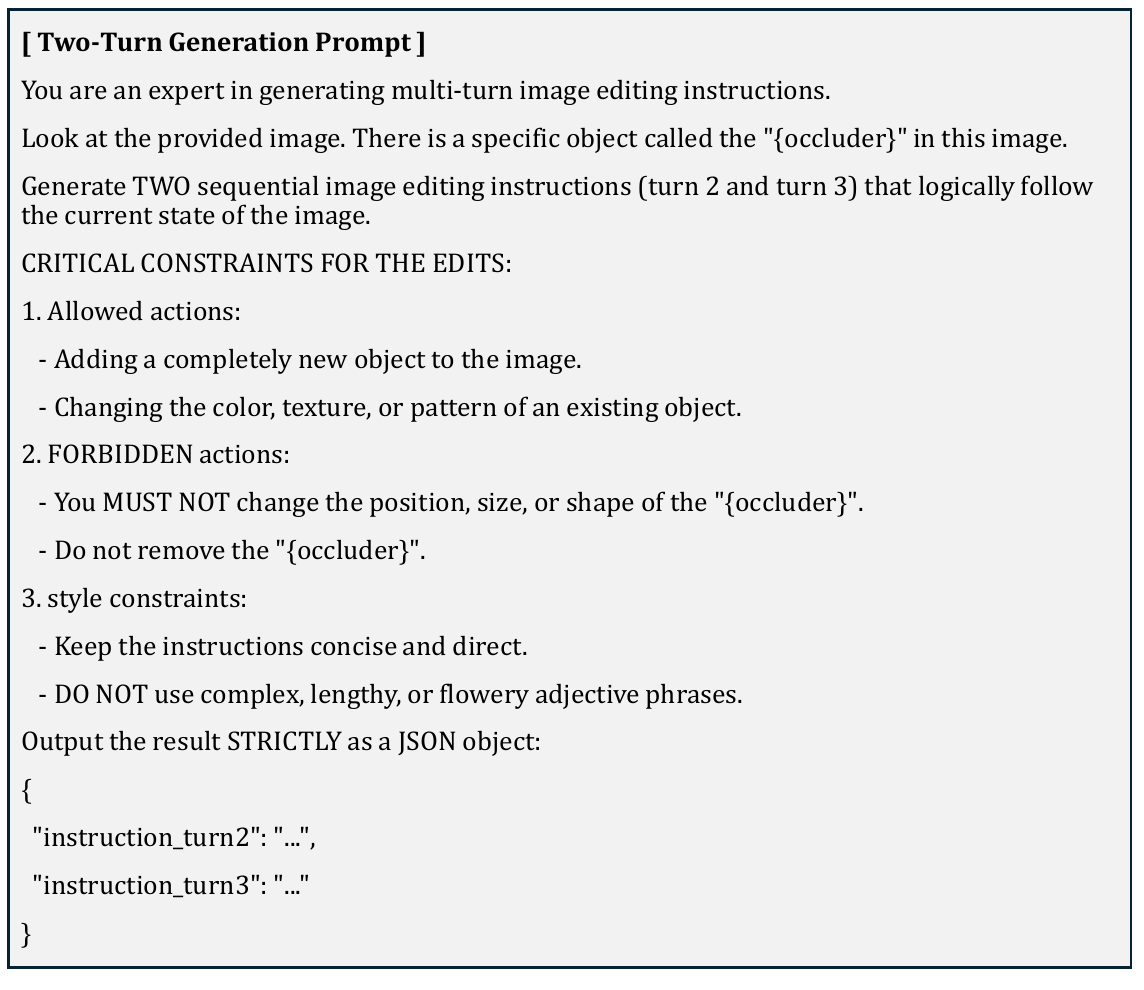}
\vspace{-0.5em}
\caption{
Two-turn generation prompt used for constructing multi-turn editing instructions.
The prompt generates two sequential edit instructions while preserving the occluder's position, size, and shape.
}
\label{fig:prompt_two_turn_generation}
\vspace{-0.5em}
\end{figure*}
\begin{figure*}[t]
\centering
\includegraphics[width=0.95\textwidth]{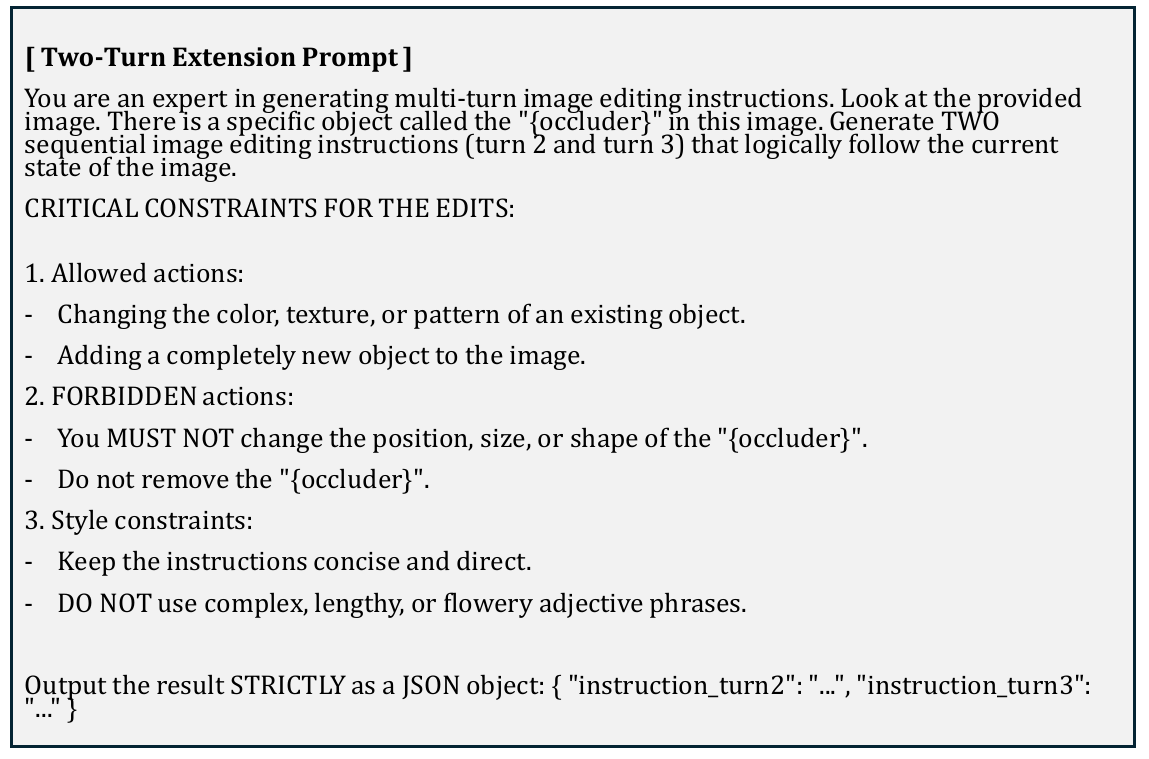}
\vspace{-0.5em}
\caption{
Two-turn extension prompt used to extend an existing editing trajectory.
The prompt adds intermediate edits while preventing changes to the occluder's position, size, or shape.
}
\label{fig:prompt_two_turn_extension}
\vspace{-0.5em}
\end{figure*}
\begin{figure*}[t]
\centering
\includegraphics[width=0.95\textwidth]{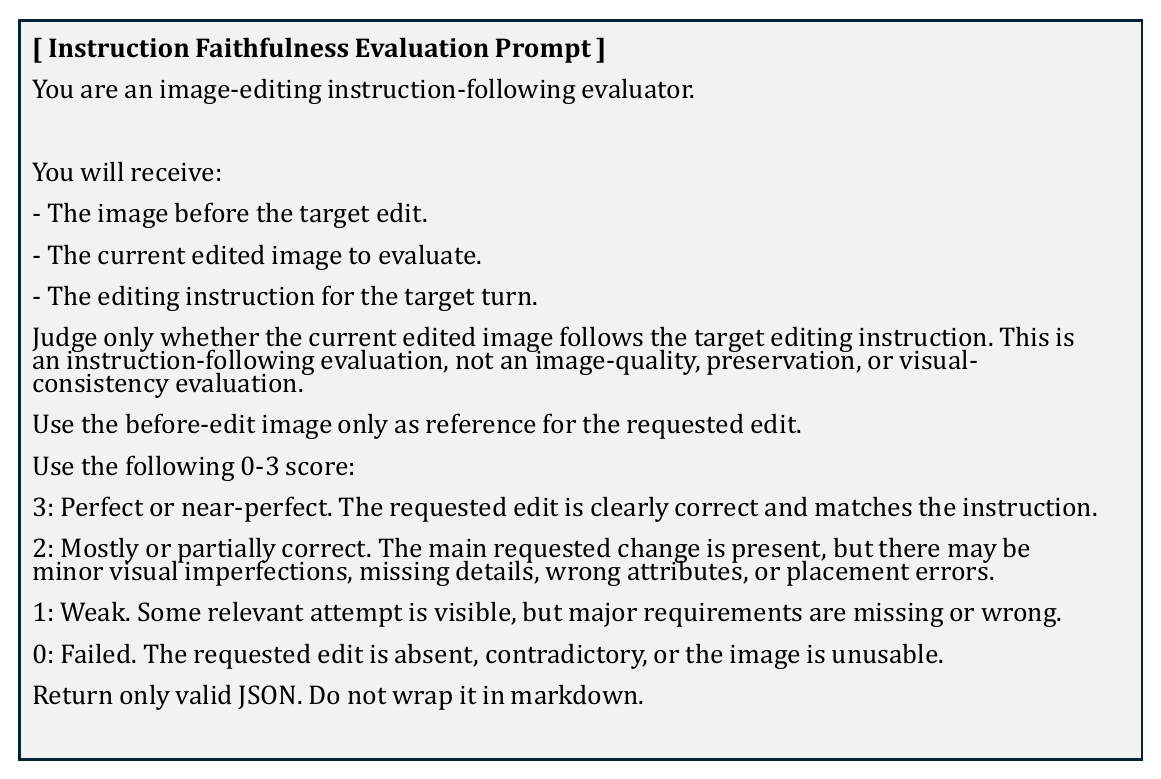}
\vspace{-0.5em}
\caption{
Instruction faithfulness evaluation prompt.
The evaluator judges only whether the edited image follows the target editing instruction, independently of image quality, preservation, or temporal consistency.
}
\label{fig:prompt_instruction_faithfulness}
\vspace{-0.5em}
\end{figure*}
\begin{figure*}[t]
\centering
\includegraphics[width=0.95\textwidth]{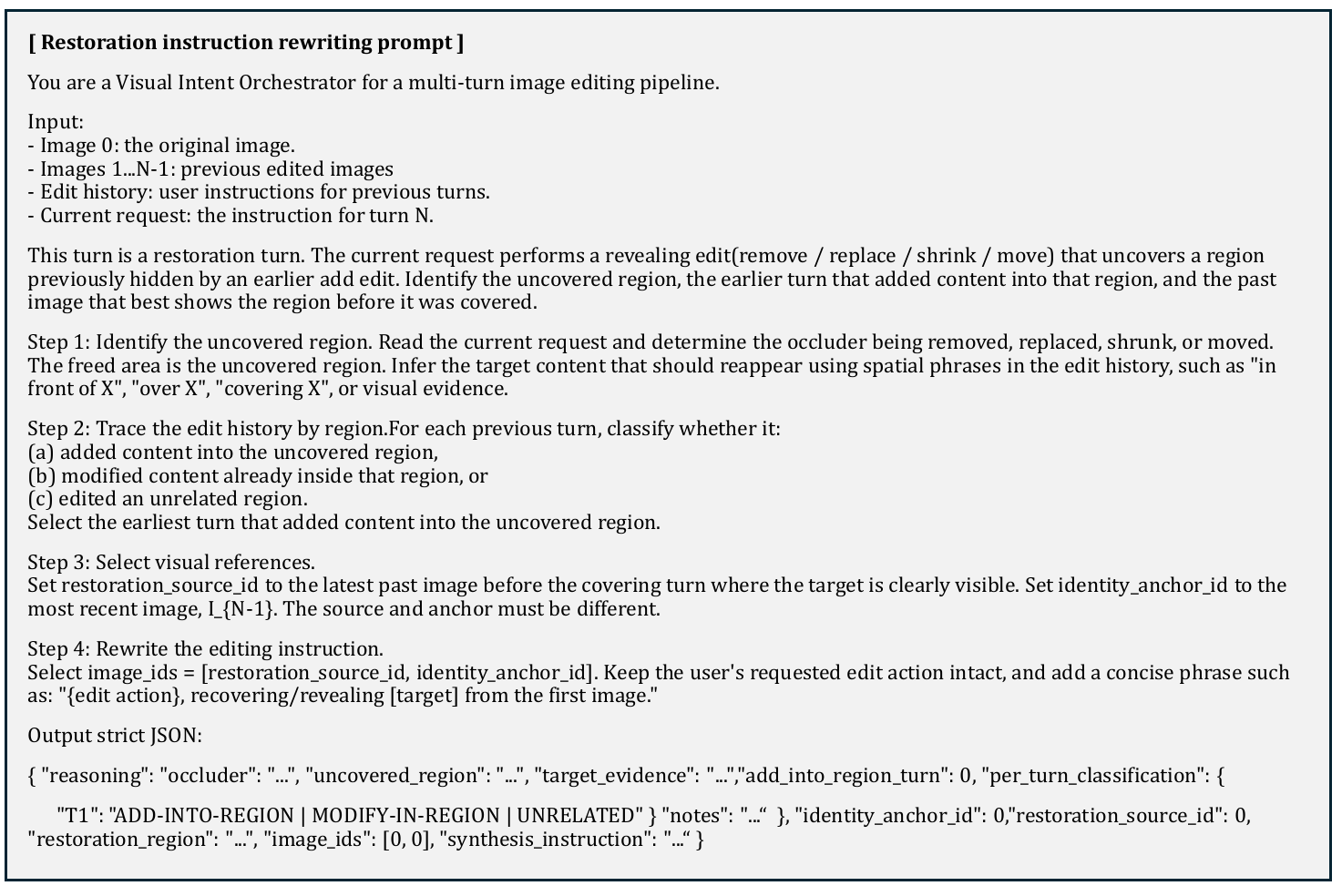}
\vspace{-0.5em}
\caption{
Restoration-aware orchestration prompt used by ReSpec.
For restoration turns, the VLM identifies the uncovered region, traces the edit history by region, selects a restoration source and identity anchor, and rewrites the user request into a restoration-aware synthesis instruction.
}
\label{fig:prompt_restore_vlm}
\vspace{-0.5em}
\end{figure*}
\begin{figure*}[t]
\centering
\includegraphics[width=0.95\textwidth]{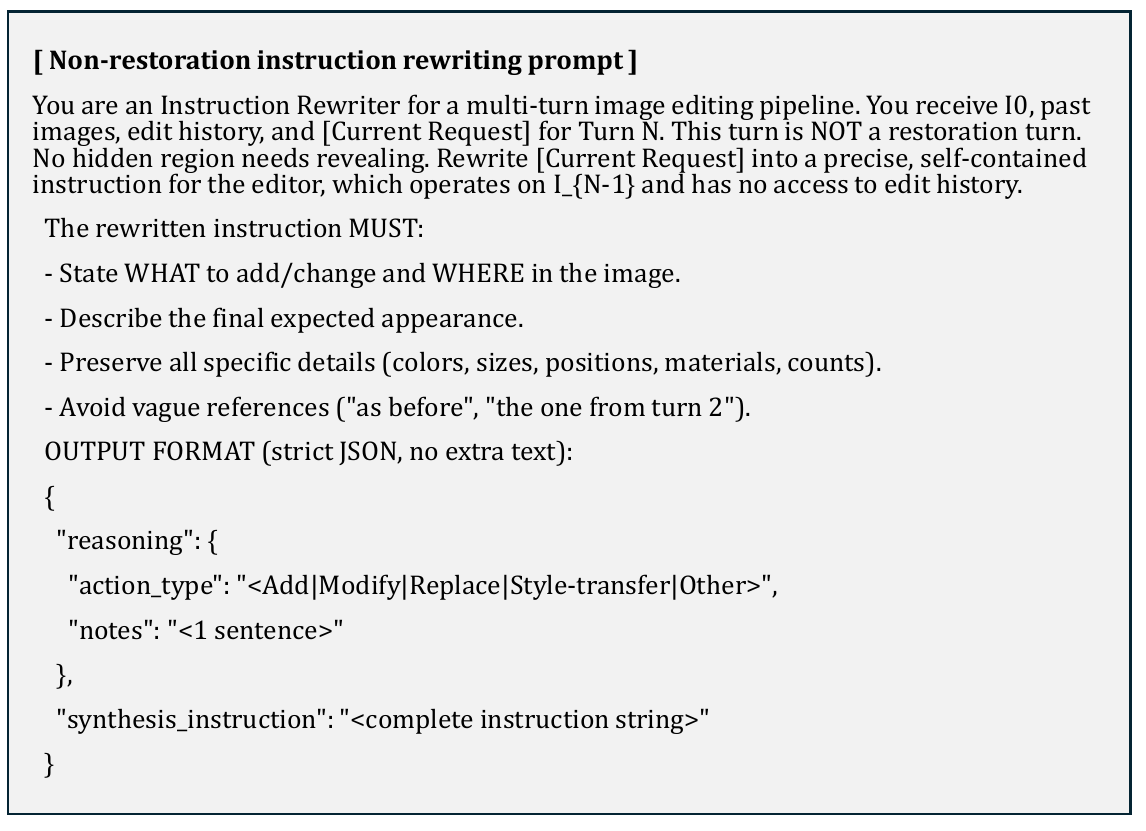}
\vspace{-0.5em}
\caption{
Non-restoration instruction rewriting prompt used by ReSpec.
For turns that do not reveal hidden content, the VLM rewrites the current request into a precise, self-contained instruction for the editor.
}
\label{fig:prompt_nonrestore_vlm}
\vspace{-0.5em}
\end{figure*}

\end{document}